\documentclass[preprint]{elsarticle}

\usepackage{hyperref}
\usepackage{tikz,pgffor}
\usepackage{times}
\usepackage{wrapfig}
\usepackage{graphicx}
\usepackage{amsmath}
\usepackage{amssymb}
\usepackage{verbatim}
\usepackage{booktabs}
\usepackage{slgloss}
\usepackage{epstopdf}
\usepackage{rotating}
\usepackage{dblfloatfix,fixltx2e}

\usetikzlibrary{calc}
\usetikzlibrary{decorations.pathreplacing}

\journal{Computer Speech and Language}
\makeatletter
\def\ps@pprintTitle{%
 \let\@oddhead\@empty
 \let\@evenhead\@empty
 \def\@oddfoot{\centerline{\thepage}}%
 \let\@evenfoot\@oddfoot}
\makeatother
\graphicspath{{figs/}}

\renewcommand{\sp}{\overline{p}}

\renewcommand{\eqref}[1]{Eq.~(\ref{#1})}

\begin{document}

\begin{frontmatter}

\title{Lexicon-Free Fingerspelling Recognition from Video: \\ Data, Models, and Signer Adaptation}

\author[1]{Taehwan Kim}
\author[2]{Jonathan Keane}
\author[1]{Weiran Wang}
\author[1]{Hao Tang}
\author[2]{Jason Riggle}
\author[1]{Gregory Shakhnarovich}
\author[2]{Diane Brentari}
\author[1]{Karen Livescu}
\address[1]{Toyota Technological Institute at Chicago \\
6045 S Kenwood Ave, Chicago, IL 60637, USA}
\address[2]{Department of Linguistics, University of Chicago \\
1115 E. 58th Street, Chicago, IL 60637, USA}

\begin{abstract}
We study the problem of recognizing video sequences of fingerspelled
letters in American Sign Language (ASL).  Fingerspelling comprises a
significant but relatively understudied part of ASL.  Recognizing fingerspelling is challenging for a number of reasons:  It involves quick, small motions
that are often highly coarticulated; it exhibits significant variation
between signers; and there has been a dearth of
continuous fingerspelling data collected.
In this work we collect and
annotate a new data set
of continuous fingerspelling videos, compare several types of
recognizers, and explore the problem of signer variation.
 Our best-performing models are segmental (semi-Markov) conditional
 random fields using deep neural network-based features.  In the
 signer-dependent setting, our recognizers achieve up to about 92\%
 letter accuracy.  The multi-signer setting is much more
 challenging, but with neural network adaptation we achieve up to 83\%
 letter accuracies in this setting.

\end{abstract}


\end{frontmatter}

\section{Introduction}
\label{sec:intro}

Sign languages are the primary means of communication for millions of Deaf people in the world.  
In the US, there are about 350,000--500,000 people for whom American Sign Language (ASL) is the primary language~\cite{Mitchell:2006}.  
While there has been extensive research over several decades on automatic recognition and analysis of spoken language, much less progress has been made for sign languages.
Both signers and non-signers would benefit from technology that improves communication between these populations and facilitates search and retrieval in sign language video.

Sign language recognition involves major challenges.  The linguistics of sign language is less well understood than that of spoken language, hampering both scientific and technological progress.
Another challenge is the high variability in the appearance of signers' bodies and the large number of degrees of freedom. 
The closely related computer vision problem of articulated pose estimation and tracking remains largely unsolved.

\begin{figure}
\centering
\includegraphics[width=\textwidth]{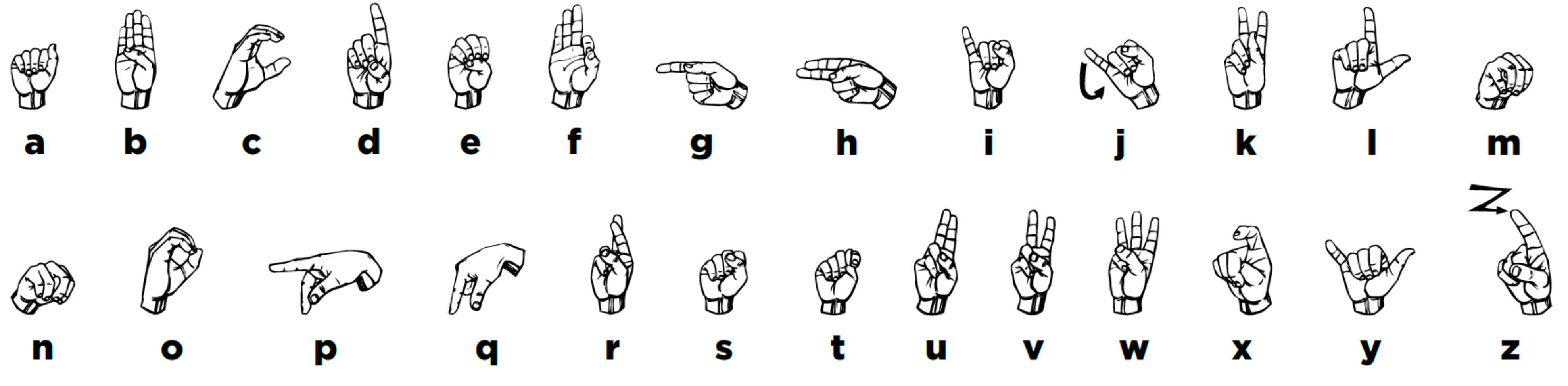}
\caption{The ASL fingerspelled alphabet.  Reproduced from~\cite{Keane2014diss}.}
\label{fig:alphabet}
\end{figure}

Signing in ASL (as in many other sign languages) involves the simultaneous use of handshape, location, movement, orientation, and non-manual behaviors.  There has now been a significant amount of research on sign language recognition, for a number of sign languages.  Prior research has focused more on the larger motions of sign and on interactions between multiple body parts, and less so on handshape.  In some contexts, however, handshape carries much of the content, and it is this aspect of ASL that we study here.

Sign language handshape has its own phonology that has been studied and enjoys a broadly agreed-upon understanding relative to the other manual parameters of movement and place of articulation~\cite{bren}.
Recent linguistic work on sign language phonology has developed approaches based on articulatory features, related to motions of parts of the hand \cite{DingM09,Keane2014diss}.
At the same time, computer vision research has studied pose estimation and tracking of hands \cite{tang2013real}, but usually not in the context of a grammar that constrains the motion.  
There is therefore a great need to better understand and model the handshape properties of sign language.

\begin{figure}[t]
\includegraphics[width=\linewidth]{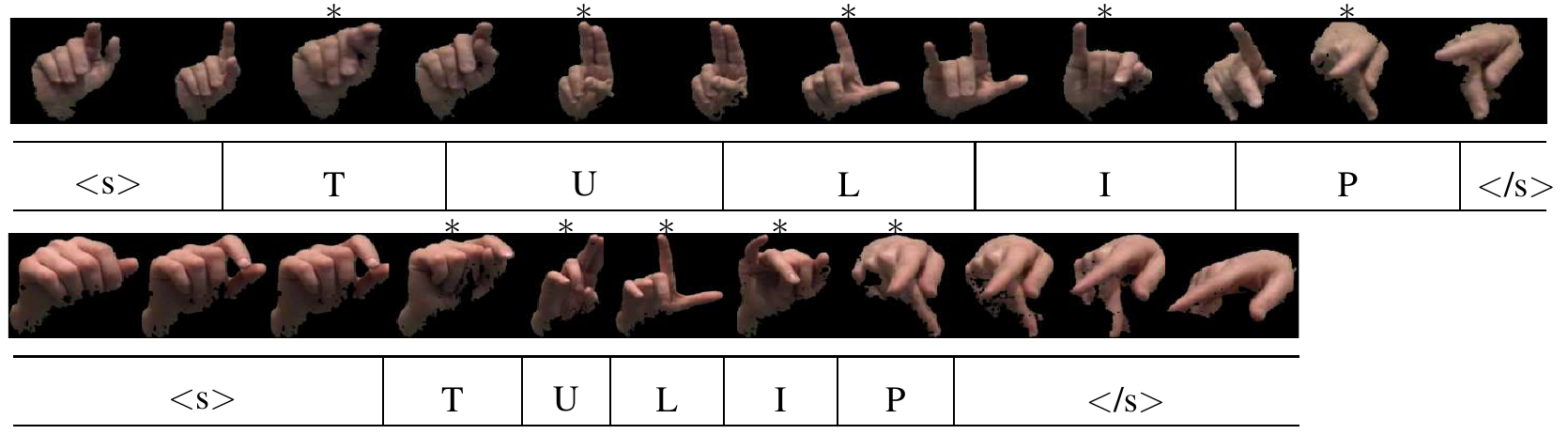}
\includegraphics[width=\linewidth]{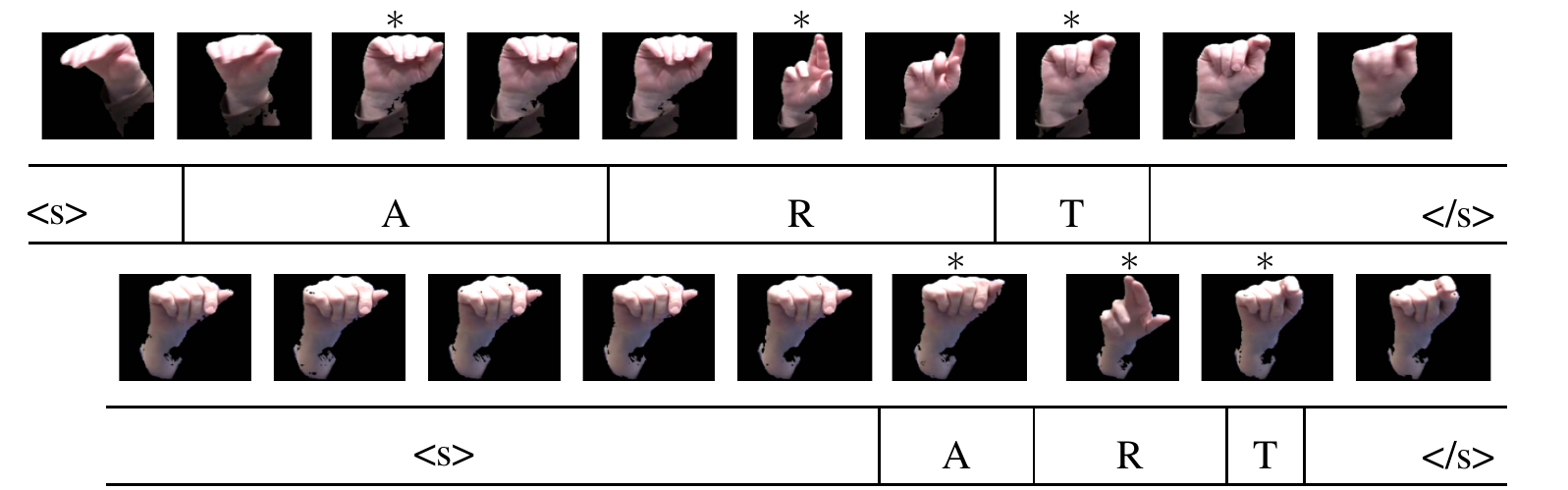}
\caption{Images and ground-truth segmentations of the fingerspelled
  words \fs{TULIP} and \fs{ART} produced by two signers in our data set. Image frames are sub-sampled
  at the same rate from both signers to show the true relative speeds.
  The starred frames indicate manually annotated peak handshapes (peak-HSs) for each FS-letter (see the text for the full definition of FS-letter and peak-HS).
  ``\texttt{<s>}'' and ``\texttt{</s>}'' denote non-signing intervals
  before/after signing.  See Sec.~\ref{sec:data},~\ref{sec:method} for more
  details on data collection, annotation, and segmentation.}
\label{fig:ex}
\end{figure}

This project focuses mainly on one constrained,
but very practical, component of ASL:  fingerspelling.  In certain contexts
(e.g., when no sign exists for a word such as a name, to introduce a
new word, or for emphasis), signers use fingerspelling: They spell out
the word as a sequence of handshapes or hand trajectories
corresponding to individual letters.  Fig.~\ref{fig:alphabet} shows
the ASL fingerspelled alphabet, and Figs.~\ref{fig:ex} show
examples of real fingerspelling sequences.  We will refer to the handshapes in Fig.~1 as fingerspelled letters (FS-letters), which are canonical target handshapes for each of the 26 letters, and the starred actual handshapes in Fig.~2 as peak handshapes.

Fingerspelled words arise naturally in the context of technical conversation or conversation about current events, such as in Deaf blogs or news sites.\footnote{E.g., \tt{http://deafvideo.tv, http://aslized.org}.}  Automatic fingerspelling recognition could add significant value to such resources.
Overall, fingerspelling comprises 12-35\% of ASL~\cite{Padden:2003}, depending on the context, and includes 72\% of the handshapes used in ASL~\cite{Brentari:2001}.  
These factors make fingerspelling an excellent testbed for handshape recognition.

Most previous work on fingerspelling and handshape has focused on
restricted conditions such as careful articulation, isolated signs,
restricted (20-100 word) vocabularies, or signer-dependent
applications~\cite{goh,Liwicki-Everingham-09,Ricco-Tomasi-09} (see
Section~\ref{sec:related_work} for more on related work).  In such
restricted settings, letter error rates (Levenshtein distances between
hypothesized and true FS-letter sequences, as a proportion of the number
of true FS-letters) of 10\% or less have been obtained.  In this work we
consider {\it lexicon-free} fingerspelling
sequences produced by
multiple signers.  This is a natural setting, since fingerspelling is
often used for names and other ``new'' terms that do not appear in any
closed vocabulary.  Our long-term goals are to develop techniques for
robust automatic detection and recognition of fingerspelled words in
video, and to enable generalization across signers, styles, and
recording conditions, both in controlled visual settings and ``in the wild.''  To this end, we are also interested in developing multi-signer, multi-style corpora of fingerspelling, as well as efficient annotation schemes for the new data.  

The work in this paper represents our first steps toward this goal:
studio collection and annotation of a new multi-signer connected
fingerspelling data set (Section~\ref{sec:data}) and high-quality
fingerspelling recognition in the signer-dependent and
multi-signer settings (Section~\ref{sec:exp}).  The new data
set, while small relative to typical speech data sets, comprises the
largest fingerspelling video data set of which we are aware containing
connected multi-signer fingerspelling that is not restricted to any particular lexicon.  Next, we compare several
recognition models inspired by automatic speech recognition, with some
custom-made characteristics for ASL
(Section~\ref{sec:method}).  We begin with a tandem hidden Markov
model (HMM) approach, where the features are based on posteriors of
deep neural network (DNN) classifiers of letter and handshape phonological
features.  We also develop discriminative segmental models, which
allow us to introduce more flexible features of fingerspelling
segments.  In the category of segmental models, we compare models for
rescoring lattices and a first-pass segmental model, and the latter
ultimately outperforms the others.  Finally, we address the problem
of signer variation via adaptation of the DNN classifiers, which
allows us to bridge much of the gap between signer-dependent and
signer-independent performance. \footnote{Parts of this
  work have appeared in our conference
  papers~\cite{kim2012,kim2013,kim2016}.  This paper includes additional model comparisons and improvements, different linguistic feature
  sets, and detailed presentation of the collected data and annotation.}

\section{Related work}
\label{sec:related_work}

Automatic sign language recognition has been approached in a variety of ways, including approaches based on computer vision techniques and ones inspired by automatic speech recognition.  Thorough surveys on the topic are provided by Koller {\it et al.}~\cite{koller2015continuous} and Ong and Ranganath~\cite{ong2005automatic}.  Here we focus on the most closely related work.

A great deal of effort has been aimed at exploiting specialized equipment such as depth sensors, motion capture, data gloves, or colored gloves in video~\cite{Oz-Leu-05,Wang-Popovic-09,pugeault2011spelling,keskin2012hand,lu2014collecting,zhang2015histogram,dong2015}.  This approach is very attractive for designing new communication interfaces for Deaf individuals.  In many settings, however, video is more practical, and for online or archival recordings is the only choice.  In this paper we restrict the discussion to the video-only setting.

A number of sign language video corpora have been collected \cite{Martinez-et-al-02, Dreuw-et-al-08, von2008significance, Dreuw-et-al-2010, Athitsos-et-al-10,neidle2012new}.  Some of the largest data collection efforts have been for European languages, such as the Dicta-Sign~\cite{efthimiou2012sign} and SignSpeak~\cite{forster2012rwth} projects.
For American Sign Language, the American Sign Language Lexicon Video Dataset (ASLLVD)~\cite{Athitsos-et-al-10,neidle2012new,ASLLVD} includes recordings of almost 3,000 isolated signs and can be searched via a query-by-example interface.  The National Center for Sign Language and Gesture Resources (NCSLGR) Corpus includes videos of continuous ASL signing, with over 10,000 sign tokens including about 1,500 fingerspelling sequences, annotated using the SignStream linguistic annotation tool~\cite{NCSLGR,Neidle_Sclaroff_Athitsos_2001}.  The latter includes the largest previous data set of fingerspelling sequences of which we are aware.  In order to explicitly study fingerspelling, however, it is helpful to have a collection that is both larger and annotated specifically with fingerspelling in mind.  To our knowledge, the data collection and annotation we report in this paper (see Sec.~\ref{sec:data}) is the largest currently available for ASL fingerspelling.

Sign language recognition from video begins with front-end features.  Prior work has used a variety of visual 
features, including ones based on estimated position, shape and movement of the hands and
head~\cite{Bowden-et-al-04,Yang-Sarkar-06,Zahedi-et-al-06,Zaki-Shaheen-11}, sometimes combined with appearance
descriptors~\cite{Farhadi-et-al-07,Dreuw-et-al-07,Nayak-et-al-09,DingM09}
and color models~\cite{Buehler11,Pfister12}.  In this work, we are
aiming at relatively small motions that are difficult to track a
priori, and therefore begin with general image appearance features
based on histograms of oriented gradients (HOG)~\cite{Dalal-Triggs-05}, which have also been used in other recent sign language recognition work~\cite{koller2015continuous,jangyodsuksign}.

Much prior work has used hidden Markov model (HMM)-based approaches~\cite{Starner-Pentland-98,Vogler-Metaxas-99,Dreuw-et-al-07,koller2015continuous}, and this is the starting point for our work as well.  Ney and colleagues have shown that it is possible to borrow many of the standard HMM-based techniques from automatic speech recognition to obtain good performance on naturalistic German Sign Language videos~\cite{Dreuw-et-al-07,koller2015continuous}.
In addition, there have been efforts involving conditional
models~\cite{Yang-Sarkar-06} and more complex (non-linear-chain)
models~\cite{ThaEtal2011,VoglerM03,theodorakis2009product}.  

As in acoustic and even visual speech recognition, the choice of basic linguistic unit is an important research question.  For acoustic speech, the most commonly used unit is the context-dependent phoneme~\cite{odell:thesis}, although other choices such as syllables, articulatory features, and automatically learned units have also been considered~\cite{Ost99,livescu2012subword}.  As the research community started to consider visual speech recognition (``lipreading''), analogous units have been explored:  visemes, articulatory features, and automatic clusters~\cite{potamianos2015audio,hazen2004segment,saenko2009multistream}.
While sign language shares some aspects of spoken language, it has some unique characteristics.  A number of linguistically motivated representations of handshape and motion have been used in some prior work, e.g.,~\cite{Bowden-et-al-04,DingM09,ThaEtal2011,VoglerM03,VoglerM01,VoglerM99,Vogler-Metaxas-99,Pit2011,The2010}, and multiple systems of phonological and phonetic features have been developed by linguists~\cite{Brentari:1998,Keane2014diss}.  One of the unique aspects of sign language is that transitional movements occupy a larger portion of the signal than steady states, and some researchers have developed approaches for explicitly modeling the transitions as units~\cite{VoglerM97,Yang2010,The2011,li2016sign}.

A subset of ASL recognition work has focused specifically on fingerspelling and/or handshape classification~\cite{Athitsos-et-al-04,Rousos-et-al-10,ThaEtal2011,pugeault2011spelling} and fingerspelling sequence recognition~\cite{goh,
Liwicki-Everingham-09,Ricco-Tomasi-09}.  Letter error rates of 10\% or less have been achieved when the recognition is constrained to a small (up to 100-word) lexicon of allowed sequences.
Our work is the first of which we are aware to address the task of {\it lexicon-free} fingerspelling
sequence recognition.

The problem of signer adaptation has been addressed in prior work, using techniques borrowed from speaker adaptation for speech recognition, such as maximum likelihood linear regression and maximum a posteriori estimation~\cite{koller2015continuous,von2008rapid,ong2004deciphering}, and to explicitly study the contrast between signer-dependent and multi-signer fingerspelling recognition.

The models we use and develop in this paper are related to prior work in both vision and speech (unrelated to sign language).  Our HMM baselines are tandem models, analogous to ones developed for speech recognition~\cite{Her00,grezl2007probabilistic}.  The segmental models we propose are related to segmental conditional random fields (SCRFs) and their variants, which have now been applied fairly widely for speech recognition~\cite{zweig,tang2015,lu2016segmental,zweig2011speech,he2012efficient}.
In natural language processing, semi-Markov CRFs have been used for named entity recognition~\cite{sarawagisemi}, where the labeling is binary.
Finally, segmental models have been applied to vision
tasks such as classification and segmentation of action sequences~\cite{shi2011human,duong2005activity} with a small set of possible activities to choose from, including work on spotting and recognition of a small vocabulary of (non-fingerspelled) signs in sign language video~\cite{Choetal2009} or instrumented data capture~\cite{kong2014towards}.
One aspect that our work shares with the speech recognition work is that we have a relatively large set of labels (26 FS-letters plus non-letter ``N/A'' labels), and widely varying lengths of segments corresponding to each label (our data includes segment durations anywhere from 2 to 40 frames), which makes the search space larger and the recognition task more difficult than in the vision and text tasks.
In prior speech recognition work, this computational difficulty has
often been addressed by adopting a lattice rescoring approach, where a
frame-based model such as an HMM system generates first-pass lattices
and a segmental model rescores them~\cite{zweig,zweig2011speech}.  We
compare this approach to an efficient first-pass segmental model~\cite{tang2015}.

\section{Data collection and annotation}
\label{sec:data}

We recorded and analyzed videos of 3 native ASL signers and 1
early learner, fingerspelling a total of 3,684 word instances.
We
annotated the video by identifying the time of peak articulation (also
known as hold, posture, or target) for each fingerspelled
letter. There were 21,453 peaks in total.
The following sections describe in detail the data collection and annotation process.

The data and annotations will be released publicly, in order to make it easy for others to study fingerspelling more extensively, and to replicate and compare against our results.

\subsection{Video recording}
The data was collected across several sessions, each including all of
the words on one word list (the lists are described in detail below). Our use of ``words'' here includes both
real words and nonsense letter sequences.
The signers were presented with an isolated word
on a computer screen. They were asked to fingerspell the word, and
then press either a green button to advance or a red button
to repeat
the word if they felt they had made a mistake. For most sessions the
signers were asked to fingerspell at a normal, natural
speed.\footnote{The instructions, given in ASL, were to:
  ``proceed at normal speed and in your natural way of
  fingerspelling.''}
Each session lasted 25--40 minutes, with a self-timed
rest/stretch break in the middle of each session.

Three word lists were created and used to collect data. The first list
had 300 words: 100 names, 100 nouns, and 100 non-English words.
These words were chosen to get examples of many letters in many contexts.
The second list
consisted of 300 mostly non-English words in an effort to get examples of each possible letter bigram. The third list
had the 300 most common nouns in the CELEX corpus in order to get a list of words that are reasonably familiar to the signers. Additionally, a set of carefully articulated, isolated fingerspelled letters were also collected from each signer.  A full listing of the word lists can be found in~\cite{Keane2014diss}.

The video was recorded in a laboratory setting, with the signers wearing green or blue clothing and with the signers set against a green background. For most sessions the signers sat in a chair with
an armrest that they could use if they felt the desire to. In a small
number of sessions the signers were asked to stand rather than
sit. Video was recorded using at least two cameras, each at an approximately 45 degree
angle from a direct frontal view.
Each camera recorded video at 1920x540 pixels per field\footnote{Through our deinterlacing process, additional lines were interpolated to give us a final result of 1920x1080 pixels per frame at 60 frames per second},
60 fields
per second, interlaced,
using the AVCHD format. For purposes of annotation, the video files were processed with
FFMPEG to deinterlace, crop, resize, and reencode them for
compatibility with the ELAN annotation
software~\cite{sloetjes2008annotation}.  For purposes of recognition
experiments, the videos were kept at full size and deinterlaced only. 

\begin{figure}[!th]
  \centering
  \begin{tabular}{cc}
    \includegraphics[width=.48\linewidth]{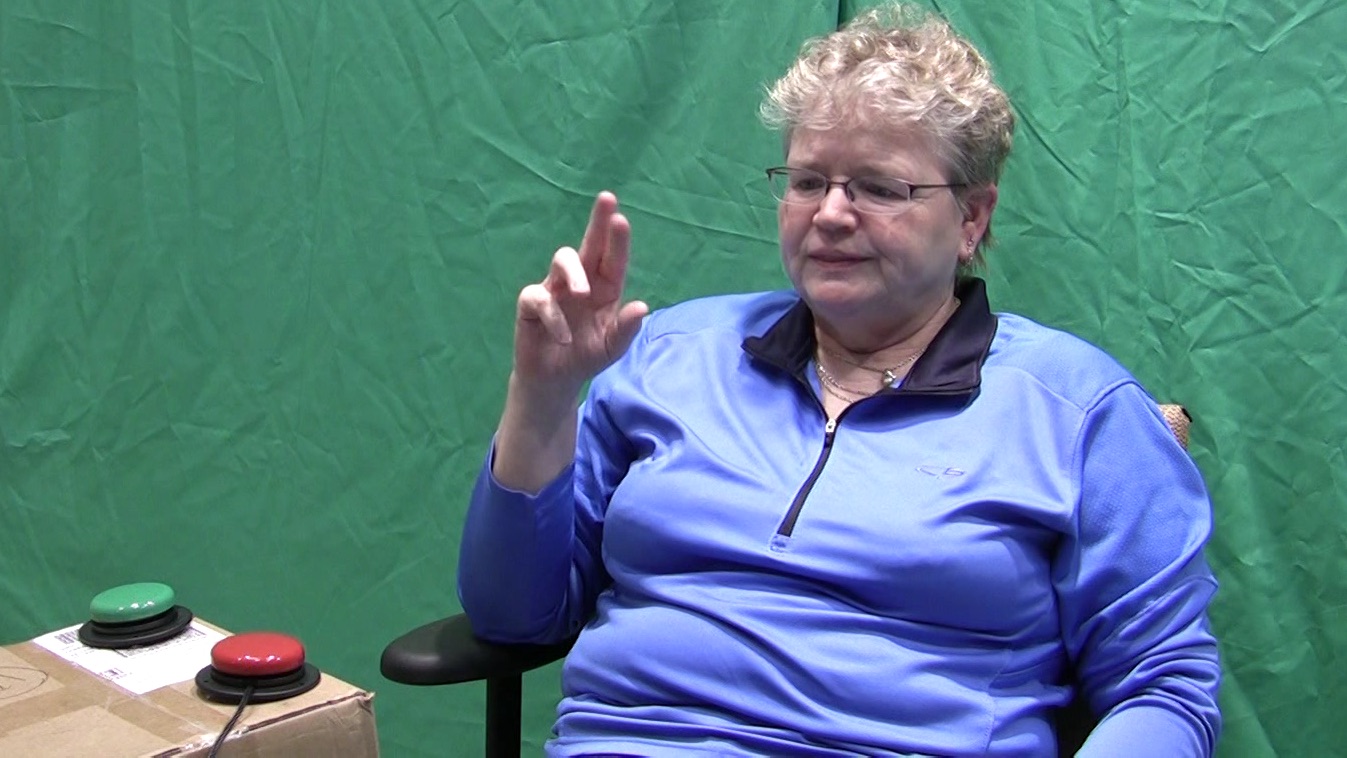}&
    \includegraphics[width=.48\linewidth]{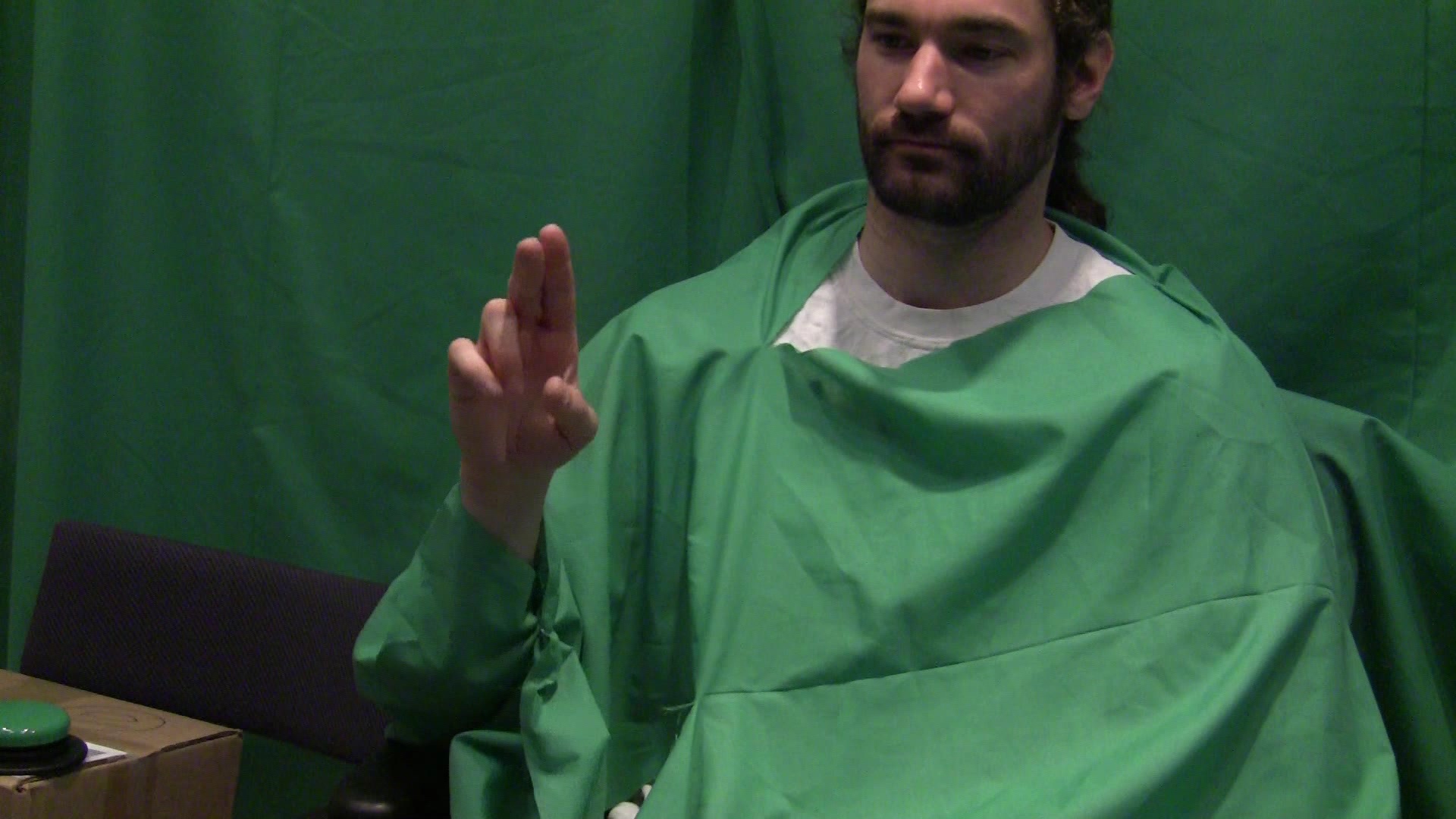}\\
    \includegraphics[width=.48\linewidth]{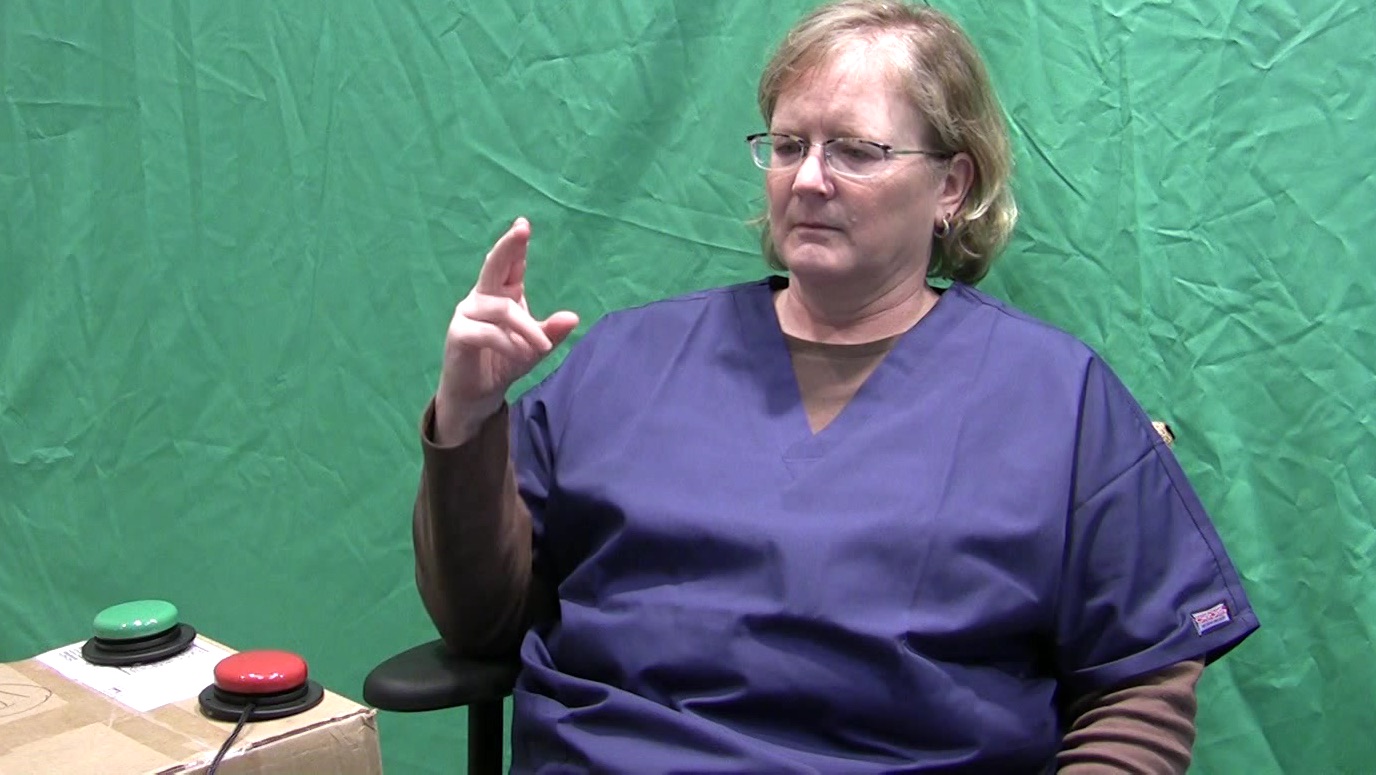}&
    \includegraphics[width=.48\linewidth]{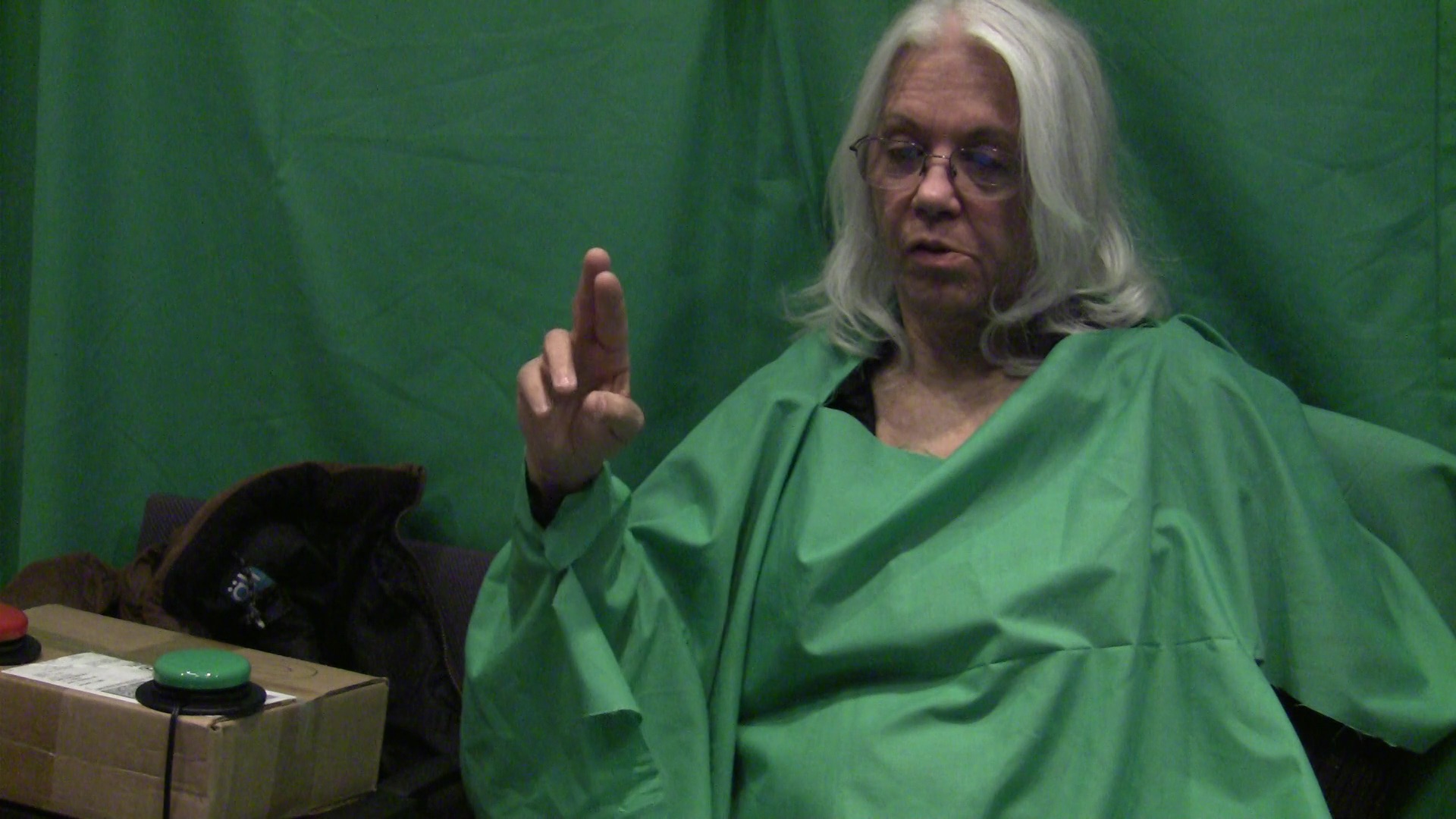}
  \end{tabular}
  \caption{Example video frames from the four signers included in the
    recognition experiments.}
  \label{fig:frames}
\end{figure}

The recording settings, including differences in environment and
camera placement across sessions, are illustrated in Fig.~\ref{fig:frames}.

\subsection{Annotation}
\label{sec:timingAnnotatoin}
Our annotation method is separated into two main parts: 1.~a simple
task to identify approximate times of each peak (\textit{peak
  detection}) and 2.~a verification task to determine precise timing
for each peak (\textit{peak verification}). The first is designed to
be extremely quick, allowing multiple annotator judgements to be
aggregated together. The second is more time-consuming and performed
by a single annotator.

\subsubsection{Peak detection}
3--4 human annotators identified the peak of each letter. For this
purpose, we defined a peak as the point where the articulators change
direction to proceed on to the next peak (i.e.~where the instantaneous
velocity of the articulators is zero or at its minimum). This point is
typically where the hand most closely resembles the canonical handshape,
although at normal speed the peak handshape is often very different
from canonical. Two FS-letters, \textsc{-j-} and
\textsc{-z-}, are not well-represented in terms of peaks, since they have
movement. For these two FS-letters, annotators were asked to
mark a peak at the point that they could determine that it was one of
these two FS-letters. Peak detection is simple, requiring
minimal training; annotators reported that this task was very
intuitive. Most of the annotators at this stage had no exposure to ASL or fingerspelling.

The peak times from the multiple annotators were averaged, after aligning
the annotations to minimize the mean absolute difference in time between the
individual annotators' peaks. We accounted for misidentified peaks by
penalizing missing or extra ones in the alignment. Using logs from
the recording session, a best guess at the FS-letter corresponding to each
peak was added by aligning the intended FS-letter sequence to the peak times (starting at the left edge of the word, matching each peak with a FS-letter).

\subsubsection{Peak verification}

Finally, a more experienced, second-language learner of ASL
or an annotator specifically trained in fingerspelling annotation
verified the location and identity of each peak from the peak
detection stage.  For the verification stage, we further refined the
definition of peak to the point in time when the handshape configuration is
closest to the canonical handshape for a given FS-letter. 
If a peak handshape remained stable for more than one frame, each stable frame was marked.
However, for recognition experiments, only the original single peak frame was used.

When using the data for recognizer training, the peak annotations are
used to segment each word into FS-letters:  The boundary between
consecutive FS-letters is defined as the midpoint between their peak frames.
\ref{sec:app} provides additional details about the
annotation and analysis conventions, as well as detailed definitions
of the fingerspelled letters.
  
\begin{figure}
\includegraphics[width=\linewidth]{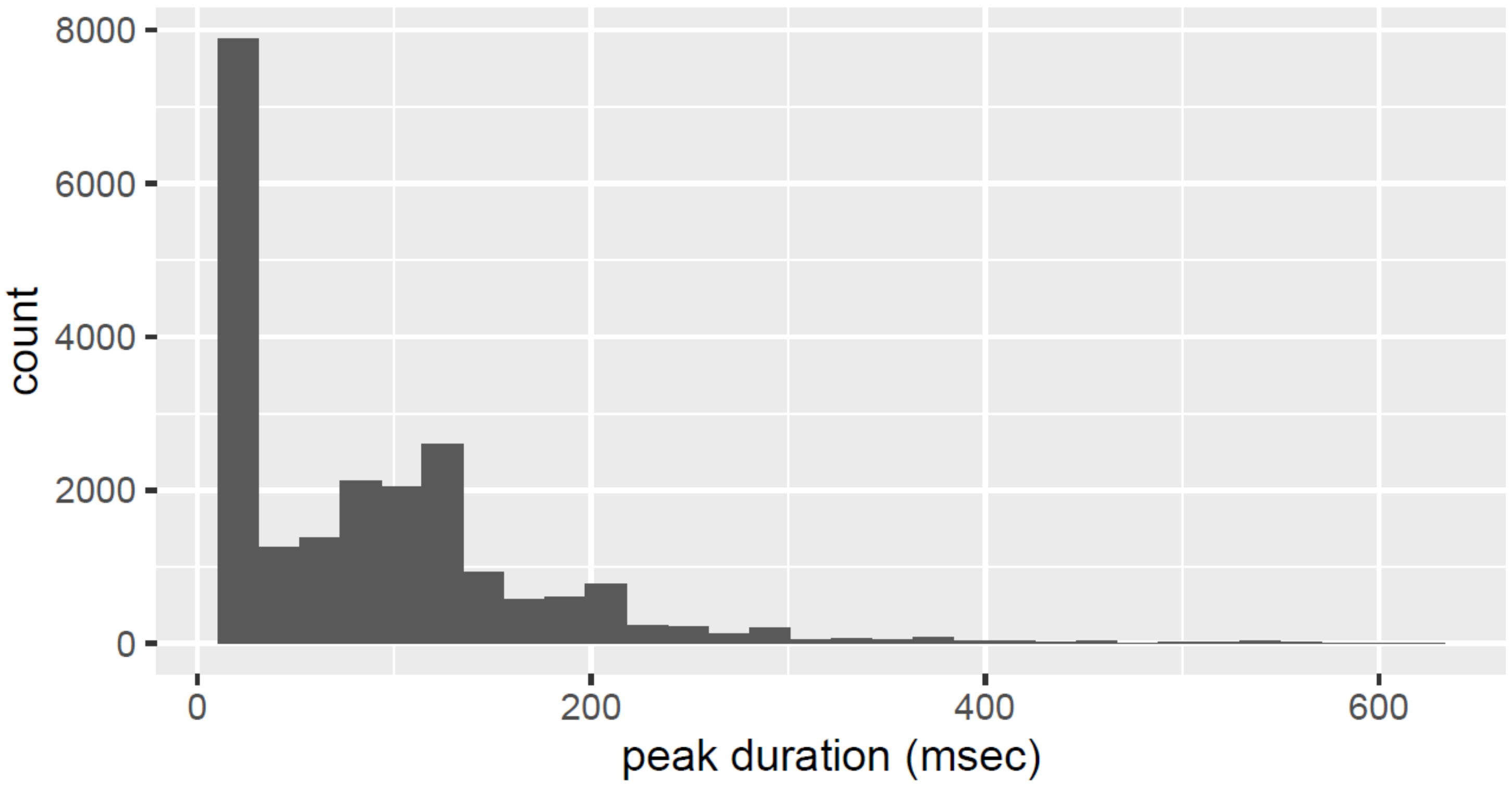}
\caption{Histogram of peak durations for all peaks in the corpus, excluding outliers. Any peak durations that were greater than 6 standard deviations more than the mean peak duration were excluded.}
\label{fig:durHistlt6sd}
\end{figure}

FS-letter frequencies in the collected data are given in Tab.~\ref{tab:freq}. A histogram of the durations of all peaks in our data (excluding outliers) is given in Fig.~\ref{fig:durHistlt6sd}.

Several studies \cite{keane_NELS, Keane2012ac, Keane2013ac, Keane2014diss} have been conducted looking at how frequently FS-letters are realized canonically (i.e.~as the correct peak handshape) in our data. The frequency of the canonical peak handshape of the FS-letter depends on a number of factors (FS-letter identity, speed, signer, etc.). Some FS-letters show up nearly always as the canonical variant (e.g.,~\fslist{c}), some almost never (\fslist{d}), while others have 75\%--85\% (\fslist[or]{e,o} respectively) canonical peak handshape realizations \cite{Keane2012ac}.  Similarly to the phonetics of spoken language, contextual effects have an impact on the phonetic configuration of each peak handshape.  For example, the pinky extension property of FS-letters has been shown to spread from peak handshapes that have an extended pinky to the ones before and after \cite{Keane2014diss}.  These coarticulation effects, combined with the quick motions of fingerspelling, are some of the factors that make the recognition task quite challenging.

\begin{table}
	\begin{centering}
		\begin{tabular}[t]{lcc}
			\hline
			peak letter  & freq & \multicolumn{1}{c}{n} \\ 
			\hline
			\nopagebreak E  & $0.1124$ & $2411$ \\
			A  & $0.0979$ & $2101$ \\
			I  & $0.0788$ & $1690$ \\
			O  & $0.0685$ & $1470$ \\
			N  & $0.0680$ & $1458$ \\
			R  & $0.0606$ & $1300$ \\
			T  & $0.0581$ & $1246$ \\
			S  & $0.0492$ & $1055$ \\
			L  & $0.0491$ & $1053$ \\
			U  & $0.0376$ & $\phantom{0}807$ \\
			C  & $0.0352$ & $\phantom{0}755$ \\
			D  & $0.0277$ & $\phantom{0}594$ \\
			M  & $0.0276$ & $\phantom{0}593$ \\
			P  & $0.0256$ & $\phantom{0}550$ \\
			\hline
		\end{tabular}
		\hspace{5mm}
		\begin{tabular}[t]{lcc}
			\hline
			peak letter  & freq & \multicolumn{1}{c}{n} \\ 
			\hline
			\nopagebreak G  & $0.0234$ & $\phantom{0}501$ \\
			H  & $0.0225$ & $\phantom{0}483$ \\
			Y  & $0.0225$ & $\phantom{0}483$ \\
			F  & $0.0201$ & $\phantom{0}432$ \\
			K  & $0.0184$ & $\phantom{0}395$ \\
			B  & $0.0170$ & $\phantom{0}365$ \\
			V  & $0.0140$ & $\phantom{0}300$ \\
			W  & $0.0135$ & $\phantom{0}290$ \\
			Z  & $0.0122$ & $\phantom{0}261$ \\
			X  & $0.0114$ & $\phantom{0}245$ \\
			J  & $0.0103$ & $\phantom{0}221$ \\
			?  & $0.0101$ & $\phantom{0}216$ \\
			Q  & $0.0083$ & $\phantom{0}178$ \\
			\hline 
		\end{tabular}
	\caption{Frequencies and counts of peaks in the corpus}
	\label{tab:freq}
\end{centering}
\end{table}

\section{Recognition methods}
\label{sec:method}

Our task is to take as input a video (a sequence of images)
corresponding to a fingerspelled word, as in Fig.~\ref{fig:ex}, and
predict the signed FS-letters.  This is a sequence prediction task
analogous to connected phone or word recognition, but there are some
interesting sign language-specific properties to the data domain.  For
example, one striking aspect of fingerspelling sequences, such as
those in Fig.~\ref{fig:ex}, is the large amount of motion and lack
of any prolonged ``steady state'' for each FS-letter.  As described in Sec.~\ref{sec:data}, each
FS-letter is represented by a brief ``peak of articulation'',
during which the hand's motion is at a minimum and the
handshape is the closest to the target handshape for the FS-letter.  This
peak is surrounded
by longer periods of motion between the current FS-letter and the
previous/next FS-letters.

Another striking property of sign language is the wide variation between signers.  Inspection of data such as Fig.~\ref{fig:ex} reveals some types of
signer variation, including differences in speed, hand appearance, and
non-signing motion before and after signing.  The speed variation is
large:  In our data, the ratio between the average per-letter duration is about $1.8$ between the fastest and slowest
signers. 

We consider
signer-dependent, signer-independent, and signer-adapted recognition.
We next describe
the recognizers we compare, as well as the techniques we explore for
signer adaptation.  All of the recognizers use deep neural network
(DNN) classifiers of FS-letters or handshape features.

\subsection{Recognizers}
In designing recognizers, we keep several considerations in mind.
First, the data set, while large in comparison to prior fingerspelling
data sets,
is still quite small compared to typical speech data sets.  This means
that large models with many context-dependent units are infeasible to
train on our data (as confirmed by our initial experiments).  We
therefore restrict attention here to ``mono-letter'' models, that is
models in which each unit is a context-independent FS-letter.  We also
consider the use of articulatory (phonological and phonetic) feature
units, and there is evidence from speech recognition research that
these may be useful in low-data
settings~\cite{stuker2003integrating,cetin}.  Second, we would
like our models to be able to capture detailed sign language-specific
information, such as the dynamic aspects of FS-letters as
discussed above; this suggests the segmental models that we consider
below.  Finally, we would like our models to be easy to adapt to new
signers.  In order to enable this, all of our recognizers use independently trained deep
neural network (DNN) classifiers, which can be adapted and plugged
into different sequence models.  Our DNNs are trained using an
L2-regularized cross-entropy loss.  The inputs are the image features
concatenated over a multi-frame window centered at the current frame, 
which are fed through several fully connected layers followed by a
softmax output layer with as many units as labels.\footnote{We have also considered convolutional neural networks with
  raw image pixels as input, but these did not significantly outperform the DNNs on image features.  See Sec.~\ref{sec:cnn-exp}.}

\subsubsection{Tandem model}
\begin{figure}[h!]
\begin{center}
        \hspace{-.5em}\includegraphics[width=0.65\linewidth]{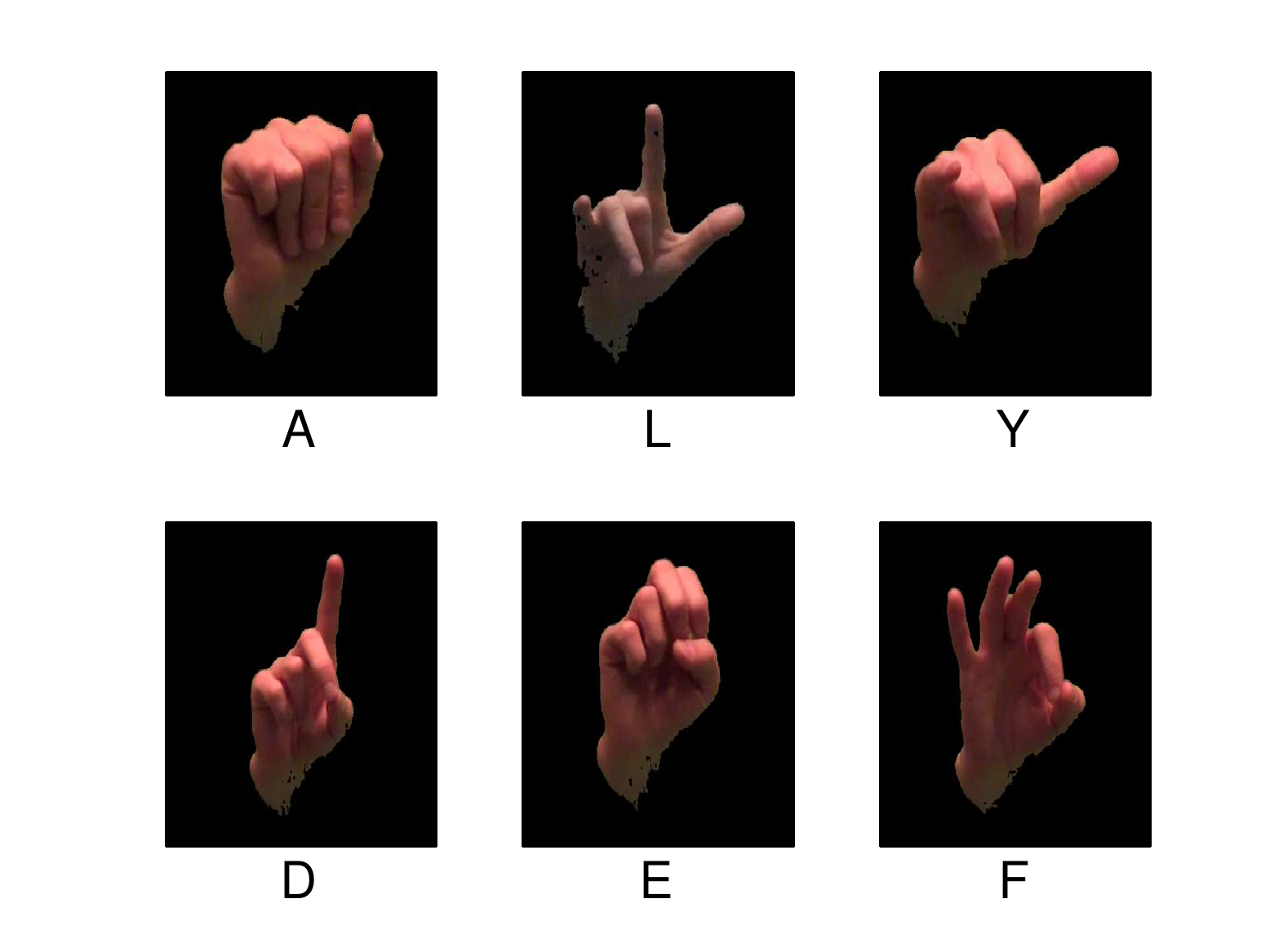}
\end{center}
\vspace{-.25in}
\caption{Example images corresponding to SF thumb = 'unopposed' (upper row) and SF thumb = 'opposed' (bottom row).}
\label{f:feature_ex}
\end{figure}

The first recognizer we consider is a fairly typical tandem model~\cite{Her00,grezl2007probabilistic}.  Frame-level
image features are fed to seven DNN classifiers, one of which predicts
the frame's FS-letter label and six others which predict handshape
phonological features.  The phonological features are defined in Tab.~\ref{t:features}, and example frames for values of one feature
are shown in Fig.~\ref{f:feature_ex}.
The classifier outputs and image features are reduced in
dimensionality via PCA and then concatenated.  The concatenated features
form the observations in an HMM-based recognizer with Gaussian mixture
observation densities.  We use a 3-state HMM for each
(context-independent) FS-letter, plus one HMM each for initial and final
``silence'' (non-signing segments).

\begin{table}[t]
\begin{center}
\begin{tabular}{|l|l|}
\hline
{\bf Feature} & {\bf Definition/Values}  \\ \hline \hline
SF point of & side of the hand where \\
reference & SFs are located  \\ \cline{2-2}
(POR) & {\it SIL, radial, ulnar, radial/ulnar} \\ \hline\hline
SF joints & degree of flexion or \\
& extension of SFs  \\  \cline{2-2}
& {\it SIL, flexed:base, flexed:nonbase,} \\ & {\it flexed:base \& nonbase,} \\
& {\it stacked, crossed, spread} \\ \hline\hline
SF quantity &combinations of SFs \\ \cline{2-2}
& {\it N/A, all, one,} \\ & {\it one $>$ all, all $>$ one} \\ \hline\hline
SF thumb & thumb position  \\ \cline{2-2}
& {\it N/A, unopposed, opposed} \\ \hline\hline
SF handpart & internal parts of the hand  \\ \cline{2-2}
& {\it SIL, base, palm, ulnar} \\ \hline\hline
UF & open/closed \\ \cline{2-2}
& {\it SIL, open, closed} \\ \hline
\end{tabular}
\caption{Definition and possible values for phonological features
  based on~\cite{bren}.  The first five features are properties of the
  active fingers ({\it selected fingers}, SF); the last feature is the
  state of the inactive or {\it unselected fingers} (UF).  In addition
  to Brentari's feature values, we add a SIL (``silence'') value to
  the features that do not have an N/A value.  For detailed
  descriptions, see~\cite{bren}.}
\label{t:features}
\end{center}
\end{table}

In addition, we use a bigram letter language model.  In general the language
model of FS-letter sequences is difficult to define or
estimate, since fingerspelling does not follow the same distribution
as English words and there is no large database
of natural fingerspelled sequences on which to train. In addition, in our
data set, the words were selected so as to maximize
coverage of letter n-grams and word types rather than
following a natural distribution.  For this work, 
the language model is trained using ARPA CSR-III text, which
includes English words and names~\cite{csriii}.  
The issue of language modeling for fingerspelling deserves more
attention in future work.

\subsubsection{Segmental CRF}
The second recognizer is a segmental CRF (SCRF).  SCRFs~\cite{sarawagisemi,zweig} are conditional log-linear models with feature functions that can be based on variable-length segments of input frames, allowing for great flexibility in defining feature functions.  
Formally, for a sequence of frames $o_1, o_2, \dots, o_T$, a segmentation of size $k$ is
a sequence of time points $0 = q_0, q_1, \dots, q_{k-1}, q_k = T$ used to denote
time boundaries of segments.  In other words, the $i$-th segment
starts at time $q_{i-1}$ and ends at $q_i$.
The labeling of segmentation $q$ is a sequence of labels
$s_1, s_2, \dots, s_k$.
We will denote the length
of label sequences and segmentations $|s|$, $|q|$, respectively.
For a segmentation of size $k$, $|s| = |q| = k$.
A SCRF defines,
over a sequence of labels $s$ given a sequence of frames $o$,
a probability distribution
\begin{equation*}
p(s | o) = \frac{\sum_{q: |q| = |s|} \exp\left(\sum_{i=1}^{|q|}
        \boldsymbol{\lambda}^\top \mathbf{f}(s_{i-1}, s_i, q_{i-1}, q_i, o)\right)}{
    \sum_{s'} \sum_{q': |q'| = |s'|} \exp\left(\sum_{i=1}^{|q'|}
        \boldsymbol{\lambda}^\top \mathbf{f}(s'_{i-1}, s'_i, q'_{i-1}, q'_i, o)\right)}
\end{equation*}
where $\boldsymbol{\lambda}$ is a weight vector, and $\mathbf{f}(s, s', q, q', o)$ is
a feature vector.
We assume $s_0$ is the sentence-start string, so that $\mathbf{f}(s_0, s_1, q_0, q_1)$ is well-defined.
In the case of sign language, it is natural to define feature functions that are sensitive to the duration and dynamics of each FS-letter segment.  

\subsubsection{Rescoring segmental CRF}
\label{sec:re_scrf}
One common way of using SCRFs is to rescore the outputs from
a baseline HMM-based recognizer, and this is one way we apply SCRFs here.
We first use the baseline recognizer, in this case the tandem
HMM, to generate lattices of high-scoring segmentations and labelings,
and then rescore them with our SCRFs.

We use the same feature functions as in~\cite{kim2013}, described here
for completeness.
Some of the feature functions are quite general to sequence
recognition tasks, while some are tailored specifically to
fingerspelling recognition.   

\paragraph{DNN classifier-based feature functions} 
The first set of feature functions measure how well the frames within
a segment match the hypothesized label.  For this purpose we use the same
DNN classifiers as in the tandem model above.

Let $y$ be a FS-letter and $v$ be the value of a FS-letter or linguistic
feature, and $g(v|o_i)$ the softmax output of a DNN
classifier at frame $i$ for class $v$.
We define several feature functions based on aggregating the DNN outputs over a segment in various ways:
\begin{itemize}
  \item $\textrm{mean}$: $f^{\text{mean}}_{yv}(s, s', q, q', o)= \delta(s'=y)\cdot\frac{1}{q' - q + 1}\sum_{i=q}^{q'} g(v|o_i)$
  \item $\textrm{max}$: $f^{\text{max}}_{yv}(s, s', q, q', o)= \delta(s'=y)\cdot \max_{i \in \{q, q + 1, \dots, q'\}} g(v|o_i)$
  \item $\textrm{div$_s$}$: a concatenation of three $\textrm{mean}$ feature functions, each computed over a third of the segment
  \item $\textrm{div$_m$}$: a concatenation of three $\textrm{max}$ feature functions, each computed over a third of the segment
\end{itemize}
These are similar to features used in prior work on SCRFs for ASR with DNN-based feature functions~\cite{abdel2013deep,he2015segmental,tang2015}.
We tune the choice of aggregated feature functions on tuning data in our experiments.

\paragraph{Peak detection features}
A sequence of fingerspelled FS-letters yields a corresponding sequence of 
peak handshapes as described above.
The peak frame and the frames around it for each FS-letter tend to have
relatively little motion, while the transition frames between peaks
have much more motion.
To encourage each predicted FS-letter segment to have a
single peak, we define letter-specific ``peak detection features''
based on approximate derivatives of the visual descriptors.  We
compute the approximate derivative as the $l_2$ norm of the difference
between descriptors in every pair of consecutive frames, smoothed by
averaging over 5-frame windows.  Typically, there
should be a single local minimum in this derivative over the span of
the segment.  We define the feature function corresponding to FS-letter
$y$ as

\vspace{-.1in}
\[f_{y}^{\textrm{peak}}(s, s', q, q', o)=\delta(s'=y)\cdot\delta_{\textrm{peak}}(o, q, q')\]

\noindent where $\delta_{\textrm{peak}}(o, q, q')$ is $1$ if there is exactly one local minimum between time point $q$ and $q'$ and 0 otherwise.

\paragraph{Language model feature} 
The language model feature is a bigram probability of the FS-letter pair corresponding to an edge:
\[
f^{\text{lm}}(s, s', q, q', o)=p_{LM}(s, s').
\]
\noindent where $p_{LM}$ is our smoothed bigram language model.

\paragraph{Baseline consistency feature}
To take advantage of the already high-quality baseline that generated
the lattices, we use a baseline feature like the one in~\cite{zweig},
which is based on the 1-best output from the baseline tandem
recognizer.  The feature has value 1 when the corresponding segment
spans exactly one FS-letter label in the baseline output and the label matches it:
\[
f^{\text{bias}}(s, s', q, q', o)=\left\{
\begin{array}{rcl}
    +1 & \text{if}~C(q, q')=1, \\
    & \phantom{\text{if}} \text{and}~B(q, q')=s' \\
    -1 & \text{otherwise} \\
\end{array}
\right.
\]
where $C(q, q')$ is the number of distinct baseline labels in the time span from $q$ to $q'$, $B(q,q')$ is the label corresponding to time span $(q,q')$ when $C(q,q') = 1$.

\subsubsection{First-pass segmental CRF}
One of the drawbacks of a rescoring approach is that the quality of the
final outputs depends both on the quality of the baseline lattices and
the fit between the segmentations in the baseline lattices and those
preferred by the second-pass model.  We therefore also consider a
first-pass segmental model, using similar features to the rescoring model.
The difference between a rescoring model and a first-pass model
lies in the set of segmentations and labellings to marginalize over.
The rescoring model only marginalizes over the hypotheses generated
by the tandem model, while the first-pass model marginalizes over
all possible hypotheses.  The first-pass model thus does not
depend on the tandem HMM.
We use a first-pass SCRF inspired by the phonetic
recognizer of Tang {\it et al.}~\cite{tang2015},
and the same feature functions as
in~\cite{tang2015}, namely average DNN outputs over each segment,
samples of DNN outputs within the segment, boundaries of DNN outputs
in each segment, duration and bias.
Recall that $g(v|o_i)$ is the softmax output of a DNN classifier
at frame $i$ for class $v$.  We use the same DNN frame
classifiers as in the rescoring setting.
Also recall that $y$ is a FS-letter and $v$ is the value of a FS-letter or linguistic feature.
The exact definition of each feature is listed below.

\paragraph{Averages of DNN outputs}
These are the same as the mean features used in the rescoring setting.

\paragraph{Samples of DNN outputs} These are samples of the DNN outputs at the mid-points
of three equal-sized sub-segments
\begin{align*}
f_{yvi}^{\text{spl}}(s, s', q, q', o) = \delta(s' = y) g(v | o_{q + (q' - q) i})
\end{align*}
for $i = 16\%, 50\%, 84\%$.

\paragraph{Left boundary} Three DNN output vectors around the left boundary of the segment
\begin{align*}
f_{yvk}^{\text{l-bndry}}(s, s', q, q', o) = \delta(s' = y) g(v | o_{q+k})
\end{align*}
for $k \in \{-1, 0, 1\}$.

\paragraph{Right boundary} Three DNN output vectors around the right boundary of the segment
\begin{align*}
f_{yvk}^{\text{r-bndry}}(s, s', q, q', o) = \delta(s' = y) g(v | o_{q'+k})
\end{align*}
for $k \in \{-1, 0, 1\}$.

\paragraph{Duration} An indicator for the duration of a segment
\begin{align*}
f_{yk}^{\text{dur}}(s, s', q, q', o) = \delta(q' - q = k) \delta(s' = y)
\end{align*}
for $k \in \{1, 2, \dots, 30\}$.

\paragraph{Bias} A constant 1
\begin{align*}
f^{\text{bias}}_y(s, s', q, q', o) = 1 \cdot \delta(s' = y).
\end{align*}

As shown in the above definitions, all features are lexicalized with the unigram label,
i.e., they are multiplied by an indicator for the hypothesized FS-letter label.

\subsection{DNN adaptation}
\label{sec:dnn}

\begin{figure*}[th]
\centering
\hspace{-.1in}\includegraphics[width=1\linewidth]{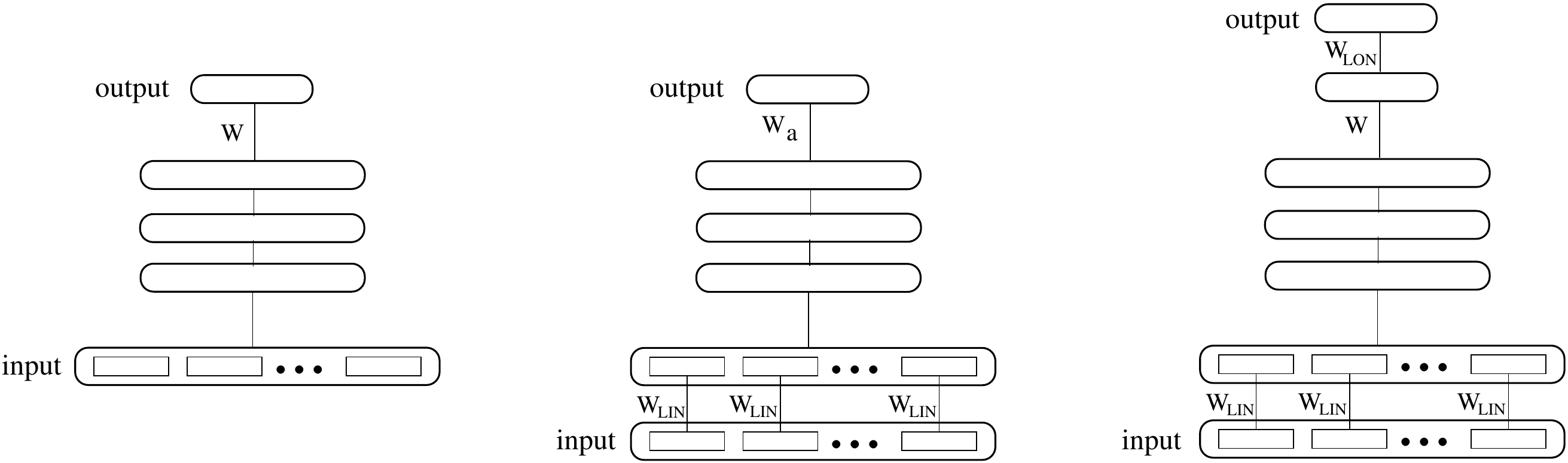}
\caption{Left:  Unadapted DNN classifier; middle:  adaptation via linear input network and output layer updating (LIN+UP); right: adaptation via linear input network and linear output network (LIN+LON).}
\label{fig:dnn-adapt}
\end{figure*}

In experiments, we will consider both signer-dependent and
signer-independent recognition.  In the latter case, the test signer
is not seen in the training set.  As we will see, there is a very large gap
between signer-dependent and signer-independent recognition on our
data.
We also consider the case where we have some labeled data from the test signer, but not a
sufficient amount for training signer-dependent models.  In this case we consider adapting a signer-independent model toward the test signer.  The most straightforward form of
adaptation for our models is to adapt only the DNN classifiers, and
then use the adapted ones in pre-trained signer-independent models.
We consider adaptation with carefully annotated adaptation data, using
ground-truth frame-level labels, as well as the setting where only
word labels are available but no frame-level alignments.  In the
latter case, we obtain frame labels via forced alignment using the signer-independent model.

Our adaptation approaches are inspired by several DNN adaptation
techniques that have been developed for speech
recognition
(e.g.,~\cite{liao2013speaker,abdel2013fast,swietojanski2014learning,doddipatla2014speaker}).
We consider several approaches, shown in
Fig.~\ref{fig:dnn-adapt}.
Two of the approaches are based on linear input networks (LIN) and
linear output networks (LON)~\cite{Neto_95a,Yao_12a,li2010}.  In these
techniques most of the network parameters are fixed to the
signer-independent ones, and a limited
set of weights at the input and/or output layers are learned. 

In the approach we refer to as LIN+UP in Fig.~\ref{fig:dnn-adapt}, we apply an affine transformation $W_{\text{LIN}}$ to the 
static features at each frame, as a pre-processing step before frame concatenation,
and use the transformed features as input to the trained signer-independent DNNs. We simultaneously learn $W_{\text{LIN}}$ and fine-tune the last (softmax) layer weights by minimizing the same cross-entropy loss
on the adaptation data, initialized with the signer-independent weights..

The second approach, referred to as LIN+LON in
Fig.~\ref{fig:dnn-adapt}, uses the same adaptation layer at the
input.  However, instead of adapting the softmax weights at the top
layer, it removes the softmax output activation and adds a new softmax
output layer $W_{\text{LON}}$ trained for the test signer.
The new input and output layers are trained jointly with the same
cross-entropy loss. 

Finally, we also consider adaptation by fine-tuning all of the DNN
weights on the adaptation data, using as initialization the
signer-independent DNN weights (that is, using the signer-independent
DNN as a ``warm start'').

\section{Experimental Results}
\label{sec:exp}

We report on experiments using the fingerspelling data from the four
ASL signers described above.  We begin by describing some of the
front-end details of hand segmentation and feature extraction,
followed by experiments with the frame-level DNN classifiers
(Sec.~\ref{sec:dnn-exp}) and FS-letter sequence recognizers (Sec.~\ref{sec:LER-exp}).\\

\noindent{\bf Hand localization and segmentation}
As in prior work~\cite{Liwicki-Everingham-09,kim2012,kim2013},
we used a simple signer-dependent model for hand detection.  
First we manually annotated hand regions in 30 frames, and we trained a mixture of Gaussians $P_{hand}$ for the color of the hand pixels in L*a*b color space, and a single-Gaussian color model $P_{bg}^x$ for every pixel $x$ in the image excluding pixel values in or near marked hand regions. 
Given the color triplet $\mathbf{c}_x=[l_x,a_x,b_x]$ at pixel $x$ from
a test frame, we assign the pixel to the hand if
\begin{equation}
  \label{eq:handodds}
  P_{hand}(\mathbf{c}_x)\pi_{hand}\,>\,
  P_{bg}^x(\mathbf{c}_x)(1-\pi_{hand}),
\end{equation}
where the prior $\pi_{hand}$ for hand size is estimated from the same 30
training frames.

We clean up the output of this simple model via several filtering
steps.  First, we suppress pixels that fall
within regions detected as faces by the Viola-Jones face detector~\cite{violajones},
since these tend to be false positives.  We also suppress pixels that
passed the log-odds test (Eq.~\ref{eq:handodds}) but have a low estimated value of $P_{hand}$, since
these tend to correspond to movements in the scene.  Finally, we suppress pixels outside of a
(generous) spatial region where the signing is expected to occur. The largest surviving connected
component of the resulting binary map is treated as a mask that
defines the detected hand region. Some examples of the resulting hand
regions are shown in
Fig.~\ref{fig:ex}. Although this approach currently involves
manual annotation for a small number of frames, it could be fully
automated in an interactive system, and may not be required if given a
larger number of training signers. \\

\noindent{\bf Handshape descriptors} We use histograms of oriented
gradients (HOG~\cite{Dalal-Triggs-05}) as the visual descriptor
(feature vector) for a given hand region.
We first resize the tight bounding box of the hand region to a
canonical size of 128$\times$128 pixels, and then compute HOG features
on a spatial pyramid of regions, 4$\times$4,
8$\times$8, and 16$\times$16 grids,  with eight orientation bins per
grid cell, resulting in 2688-dimensional descriptors.  Pixels outside of the
hand mask are ignored in this computation. For the HMM-based recognizers, to speed
up computation, these descriptors were projected to at most 200
principal dimensions; the 
exact dimensionality in each experiment was tuned on a development set. For DNN frame classifiers, we found that finer grids did not improve much with increasing complexities, so we use 128-dimensional descriptors.\\

\subsection{DNN frame classification performance}
\label{sec:dnn-exp}
\begin{figure}\centering
\centering
\vspace{-.06in}
\hspace{-.0in}\includegraphics[width=0.6\linewidth]{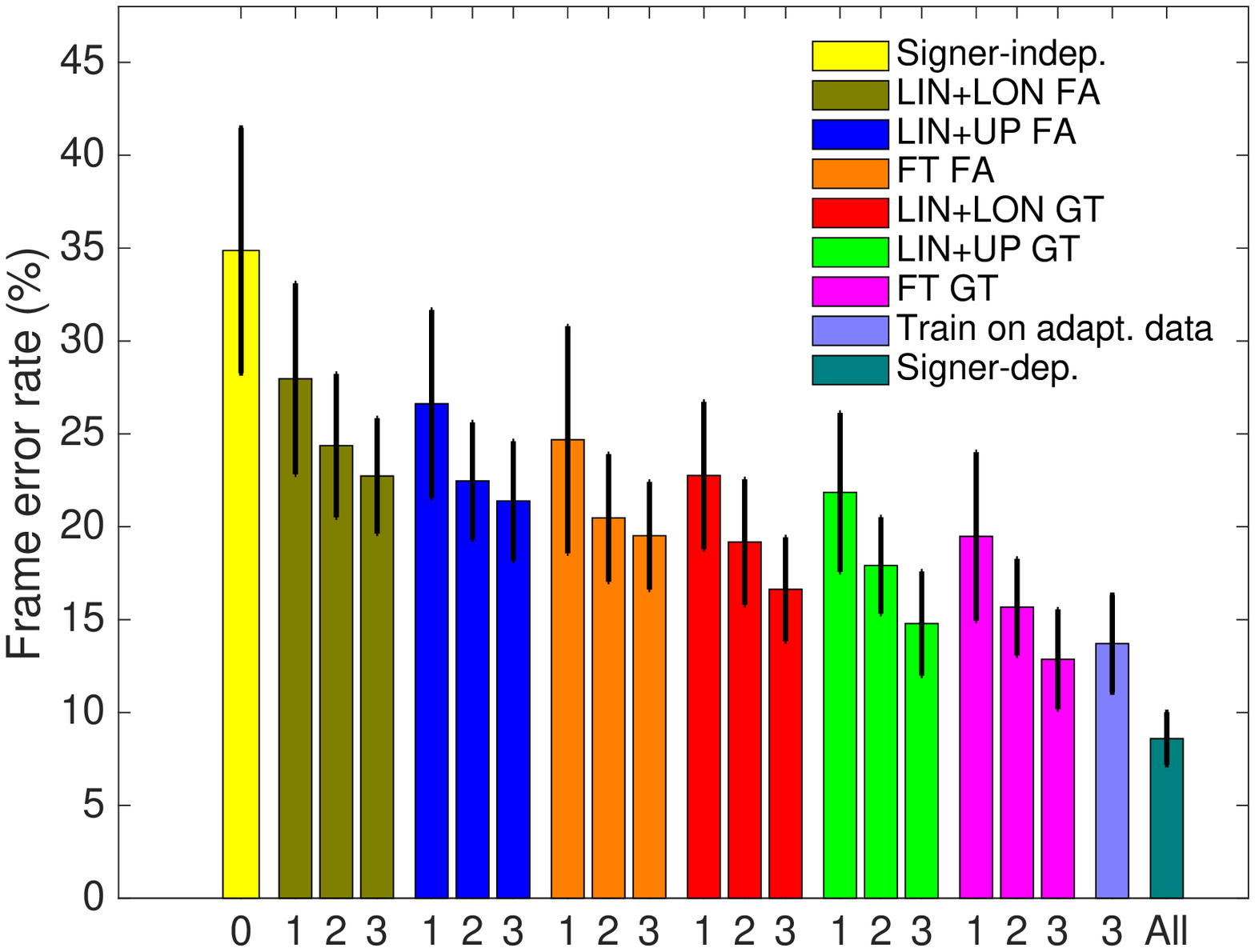} \\
\hspace{-.0in}\includegraphics[width=0.6\linewidth]{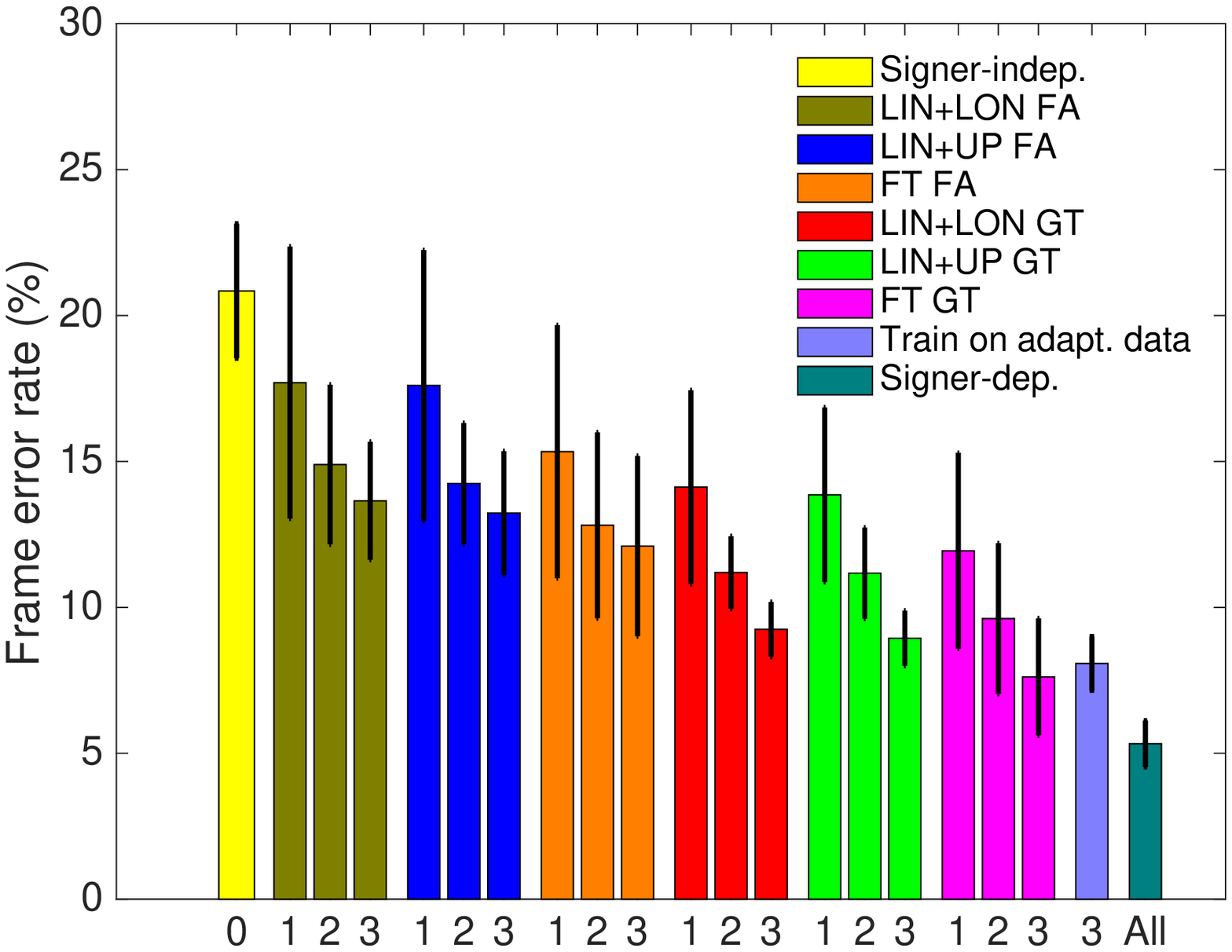}
\vspace{-.15in}
\caption[Frame errors with DNN classifiers, with various settings]{Frame errors with DNN classifiers, with various settings. The horizontal axis labels indicate the amount of adaptation data (0, 1, 2, 3 = none, 5\%, 10\%, 20\% of the test signer's data, corresponding to no adaptation (signer-independent), $\sim29$, $\sim58$, and $\sim115$ words).  GT = ground truth labels; FA = forced alignment labels; FT = fine-tuning. We also added trained DNN on only 20\% of the test signer's data. Signer-dependent DNN uses 80\% of the test signer's data.}
\label{fig:adapt_fer}
\end{figure}

Since all of our fingerspelling recognition models use DNN frame classifiers as a building block,
we first examine the performance of the frame classifiers.

For training and tuning the DNNs, we use the recognizer training data, split into 90\% for DNN training and 10\% for DNN tuning.
The DNNs are trained for seven tasks (FS-letter classification and
classification of each of the six phonological features). The input is
the $128$-dimensional HOG features concatenated over a $21$-frame window.
The DNNs have
 three hidden layers, each with $3000$ ReLUs~\cite{Zeiler_13a}.
 Network learning is done with cross-entropy training with a weight
 decay penalty of $10^{-5}$, via stochastic gradient descent (SGD)
 over $100$-sample minibatches for up to $30$ epochs, with
 dropout~\cite{Srivas_14a} at a rate of $0.5$ at each hidden layer,
 fixed momentum of $0.95$, and initial learning rate of $0.01$, which
 is halved when held-out accuracy stops improving.  We pick the best-performing epoch on held-out data.  The network
 structure and hyperparameters were tuned on held-out (signer-independent) data in initial experiments.  

We consider the signer-dependent setting (where the DNN is trained on
data from the test signer), signer-independent setting (where the DNN
is trained on data from all except the test signer), and
signer-adapted setting (where the signer-independent DNNs are adapted
using adaptation data from the test signer).
 For LIN+UP and LIN+LON, we adapt by running SGD over minibatches of $100$ samples with a fixed momentum of $0.9$ for up to $20$ epochs, with initial learning rate of $0.02$ (which is halved when accuracy stops improving on the adaptation data).  For fine-tuning, we use the same SGD procedure as for the signer-independent DNNs.  We pick the epoch with the highest accuracy on the adaptation data.

The frame error rates for all settings are given in
Fig.~\ref{fig:adapt_fer}.  For the signer-adapted case, we consider DNN
adaptation with different types and amounts of supervision.  The types
of supervision include fully labeled adaptation data (``GT'', for
``ground truth'', in the figure), where the peak locations for all
FS-letters are manually annotated; as well as adaptation data labeled
only with the FS-letter sequence but not the timing information.  In the
latter case, we use the baseline tandem system to generate forced
alignments (``FA'' in the figure).  We consider amounts of adaptation
data from 5\% to 20\% of the test signer's full data.

These results show that among the adaptation methods, LIN+UP slightly outperforms LIN+LON, and fine-tuning outperforms both
LIN+UP and LIN+LON.  For FS-letter sequence recognition experiments in
the next section, we adapt via fine-tuning using $20\%$ of the test signer's data.

We have analyzed the types of errors made by the DNN classifiers.  One
of the main effects is that all of the signer-independent classifiers
have a large number of incorrect predictions of the non-signing
classes ($<$s$>$, $<$/s$>$).  
This may be due to the previously mentioned observation that
non-linguistic gestures are variable and easy to confuse with signing
when given a new signer's image frames.  
As the DNNs are adapted, this is the main type of error that is corrected.
Specifically, of the frames with label $<$s$>$ that are misrecognized by the signer-independent DNNs, 18.3\% are corrected after adaptation; of the misrecognized frames labeled $<$/s$>$, 11.2\% are corrected.

\subsection{FS-letter recognition experiments}
\label{sec:LER-exp}

\subsubsection{Signer-dependent recognition}
Our first continuous FS-letter recognition experiments are signer-dependent; that is, we train and test on the same signer, for each of four signers. 
For each signer, we use a 10-fold setup:  In each fold, 80\% of the data is used as a training set, 10\% as a development set for tuning parameters, and the remaining 10\% as a final test set.    
We independently tune the parameters in each fold.  To make the results comparable to the later adaptation results, we use 8 out of 10 folds to compute the final test results and report the average letter error rate (LER) over those 8 folds.
For language models, we train letter bigram language models from large online dictionaries of varying sizes that include both English words and names~\cite{csr}. 
We use HTK~\cite{htk} to implement the baseline HMM-based recognizers and SRILM~\cite{srilm} to train the language models. 
The HMM parameters (number of Gaussians per state, size of language model vocabulary, transition penalty and language model weight), as well as the dimensionality of the HOG descriptor input and HOG depth, were tuned to minimize development set letter error rates for the baseline HMM system.  The front-end and language model hyperparameters were then kept fixed for the experiments with SCRFs (in this sense the SCRFs are slightly disadvantaged).   
Additional parameters tuned for the SCRF rescoring models included the N-best list sizes, type of feature functions, choice of language models, and L1 and L2 regularization parameters.  Finally, for the first-pass SCRF, we tuned step size, maximum length of segmentations, and number of training epochs. 

Tab.~\ref{t:LER} (last row) shows the signer-dependent FS-letter recognition results. SCRF rescoring improves over the tandem HMM, and the first-pass SCRF outperforms both.

Note that in our experimental setup, there is some overlap of word
types between training and test data.  This is a realistic setup,
since in real applications some of the test words will have been
previously seen and some will be new.  However, for comparison, we
have also conducted the same experiments while keeping the training,
development, and test vocabularies disjoint; in this modified setup,
letter error rates increase by about 2-3\% overall, but the SCRFs
still outperform the other models.

\begin{table*}[ht]
\centering
\resizebox{\linewidth}{!}{
\begin{tabular}{|l||r|r|r|r|r||r|r|r|r|r||r|r|r|r|r|}
\hline
        & \multicolumn{5}{c||}{Tandem HMM} & \multicolumn{5}{c||}{Rescoring SCRF} & \multicolumn{5}{c|}{First-pass SCRF} \\ \hline
Signer  & 1 & 2 &  3 &  4 & {\bf Mean} &  1 &  2 &  3 &  4 & {\bf Mean} & 1 &  2 &  3 &  4 & {\bf Mean} \\ \hline\hline
No adapt. & 54.1&	54.7&	62.6&	57.5&	{\bf 57.2}&	52.6&	51.2&	61.1&	56.3&	{\bf 55.3}&	55.3&	53.3&	72.5&	61.4&	{\bf 60.6} \\ \hline
Forced align. & 30.2&    38.5&   39.6&   36.1&   {\bf 33.6}&     39.5&   36.0&   38.2 &   34.5&   {\bf 32.0}&    24.4&   24.9&   36.5&   35.5&   {\bf 30.3} \\ \hline
Ground truth  & 22.0&    13.0&   31.6&   21.4&   {\bf 22.0}&     22.4&   13.5&   29.5&   21.4&   {\bf 21.7}&     15.2&   10.6&   24.9&   18.4&   {\bf 17.3} \\ \hline
Signer-dep. &     13.8&    7.1&   26.1&   11.5 & {\bf 14.6}& 10.2 &   7.0&   19.1&    10.0 &{\bf 11.5} & 8.1&    7.7&    9.3&   10.1&  {\bf 8.8} \\ \hline 
\end{tabular}
}
\vspace{-.1in}
\caption{Letter error rates (\%) on four test signers.}
\label{t:LER}
\end{table*}

\subsubsection{Signer-independent recognition}
\label{sec:sigindep_exp}

In the signer-independent setting, we would like to recognize
FS-letter sequences from a new signer, given a model
trained only on data from other signers.  For each of the four test
signers, we train models on the remaining three signers, and report
the performance for each test signer and averaged over the four test
signers.  For direct comparison with the signer-dependent experiments,
each test signer's performance is itself an average over the 8 test
folds for that signer.

As shown in the first line of Tab.~\ref{t:LER}, the signer-independent
performance of the three types of recognizers is quite poor, with the
rescoring SCRF somewhat outperforming the tandem HMM and first-pass
SCRF.  The poor performance is perhaps to be expected with such a
small number of training signers.

\subsubsection{Signer-adapted recognition}
\label{sec:sigadap_exp}

The remainder of Tab.~\ref{t:LER} (second and third rows) gives the connected FS-letter
recognition performance obtained with the three types of models
using DNNs adapted via fine-tuning, using different types of
adaptation data (ground-truth, GT, vs.~forced-aligned, FA).  For all
models, we do not retrain the models with the adapted DNNs, but tune
hyperparameters\footnote{See~\cite{kim2012,kim2013,tang2015}
  for details of the tuning parameters.} on 10\% of the test signer's data.  The tuned models
are evaluated on an unseen 10\% of the test signer's remaining data;
finally, we repeat this for eight choices of tuning and test sets,
covering the 80\% of the test signer's data that we do not use for
adaptation, and report the mean FS-letter accuracy over the test sets.

\begin{figure}\centering
\centering
\includegraphics[width=1.0\linewidth]{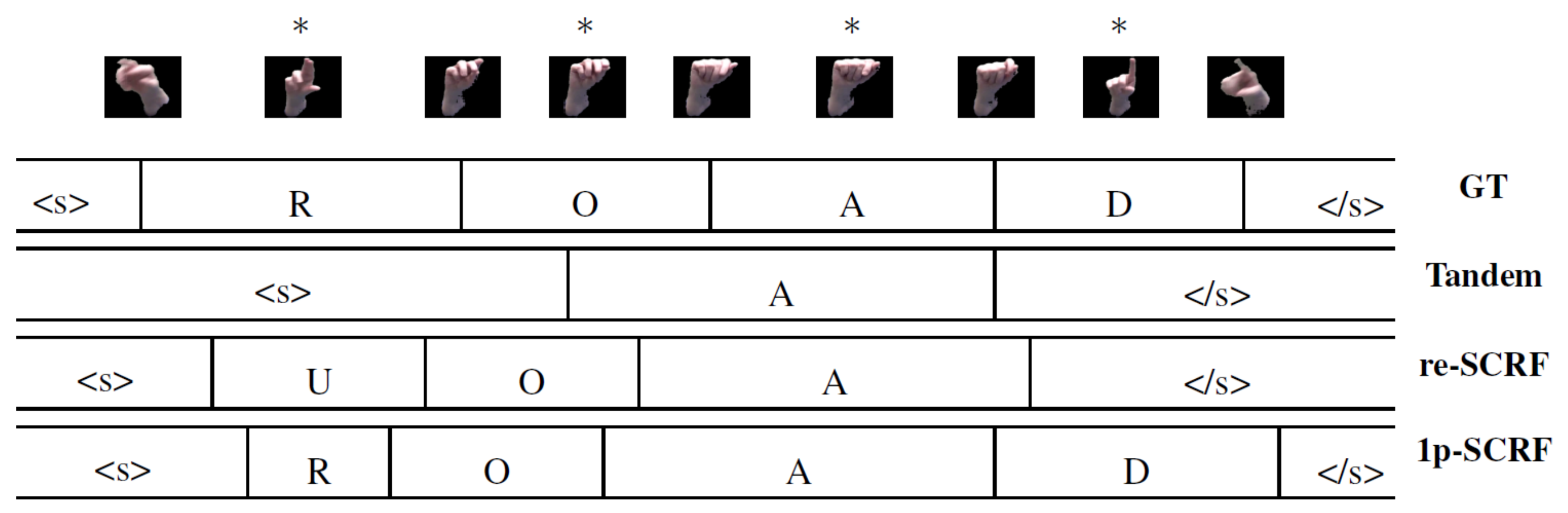}
\caption[Example frame sequence with ground-truth and hypothesized labels and segmentations.]{Sample frames from the word ROAD, with asterisks denoting the peak frame for each FS-letter and ``\texttt{$<$s$>$}'' and ``\texttt{$<$/s$>$}'' denoting periods before the first FS-letter and after the last FS-letter; ground-truth labeling and segmentation based on peak annotations (GT); hypothesized output from the tandem HMM (Tandem), rescoring SCRF (re-SCRF), and first-pass SCRF (1p-SCRF).}
\label{f:ROAD}
\end{figure}

As shown in Tab.~\ref{t:LER}, adaptation improves the performance
to up to 30.3\% letter error rate with forced-alignment adaptation
labels and up to 17.3\% letter error rate with ground-truth adaptation labels.
All of the adapted models improve similarly.  However, interestingly, the
first-pass SCRF is slightly worse than the others before adaptation
but better (by 4.4\% absolute) after ground-truth adaptation.  One
hypothesis is that the first-pass SCRF is more dependent on the DNN
performance, while the tandem model uses the original image features
and the rescoring SCRF uses that tandem model's hypotheses.
Once the DNNs are adapted, however, the first-pass SCRF outperforms
the other models.  Fig.~\ref{f:ROAD} shows an example fingerspelling sequence and the hypotheses of the tandem, rescoring SCRF, and first-pass SCRF.

\subsection{Extensions and analysis}

We next analyze our results and consider potential extensions for
improving the models.

\subsubsection{Analysis: Could we do better by training entirely on adaptation data?}
\label{sec:scratch}

Until now we have considered adapting the DNNs while using sequence models (HMMs/SCRFs) trained
only on signer-independent data.
In this section we consider alternatives to this adaptation setting.
We fix the model to a first-pass
SCRF and the adaptation data to 20\% of the test signer's data
annotated with ground-truth labels.  In this setting, we consider
two alternative ways of using the adaptation data:  (1) using the adaptation data from the
test signer to train both the DNNs and sequence model
from scratch, ignoring the signer-independent training set; and (2)
training the DNNs from scratch on the adaptation data, but using the
SCRF trained on the training signers. We compare these options with
our best results using the signer-independent SCRF and DNNs fine-tuned
on the adaptation data.  The results are shown in Tab.~\ref{t:LER_gt}.

We find that ignoring the signer-independent training set and training
both DNNs and SCRFs from scratch on the test signer
(option (1) above) works remarkably well, better than the
signer-independent models and even better than adaptation via
forced alignment (see Tab.~\ref{t:LER}).
However, training the
SCRF on the training signers but DNNs from scratch on the adaptation
data (option (2) above) improves performance further.  However,
neither of these outperforms our previous best approach of
signer-independent SCRFs plus DNNs fine-tuned on the adaptation data.

\begin{table*}[ht!]
\centering
\resizebox{\linewidth}{!}{
\begin{tabular}{|c|c|c|c|c|c||c|c|c|c|c||c|c|c|c|c|}
\hline
& \multicolumn{5}{|c||}{SCRF + DNNs trained from scratch} & \multicolumn{5}{c||}{Sig.-indep. SCRF, DNNs from scratch} & \multicolumn{5}{c|}{Sig.-indep. SCRF, fine-tuned DNN} \\ \hline
Signer & 1 & 2 & 3 & 4 & {\bf Mean} & 1 & 2 & 3 & 4 & {\bf Mean} & 1 & 2 & 3 & 4 & {\bf Mean} \\ \hline\hline
Accuracy & 23.2 &  18.2 &  28.9 &  30.1 & {\bf 25.1} & 18.4 &  12.6 &  27.0 &  20.7 & {\bf 19.7}  & 15.2&   10.6&   24.9&   18.4&   {\bf 17.3} \\ \hline
\end{tabular}
}
\vspace{-.1in}
\caption{Letter error rates (\%) for different settings of SCRF and DNN
  training in the signer-adapted case.  Details are given in Section~\ref{sec:scratch}.}
\label{t:LER_gt}
\end{table*}

\subsubsection{Analysis:  FS-letter vs.~feature DNNs}

\begin{figure*}[h!]
\centering
\begin{tabular}{cc}
  \includegraphics[width=0.8\linewidth]{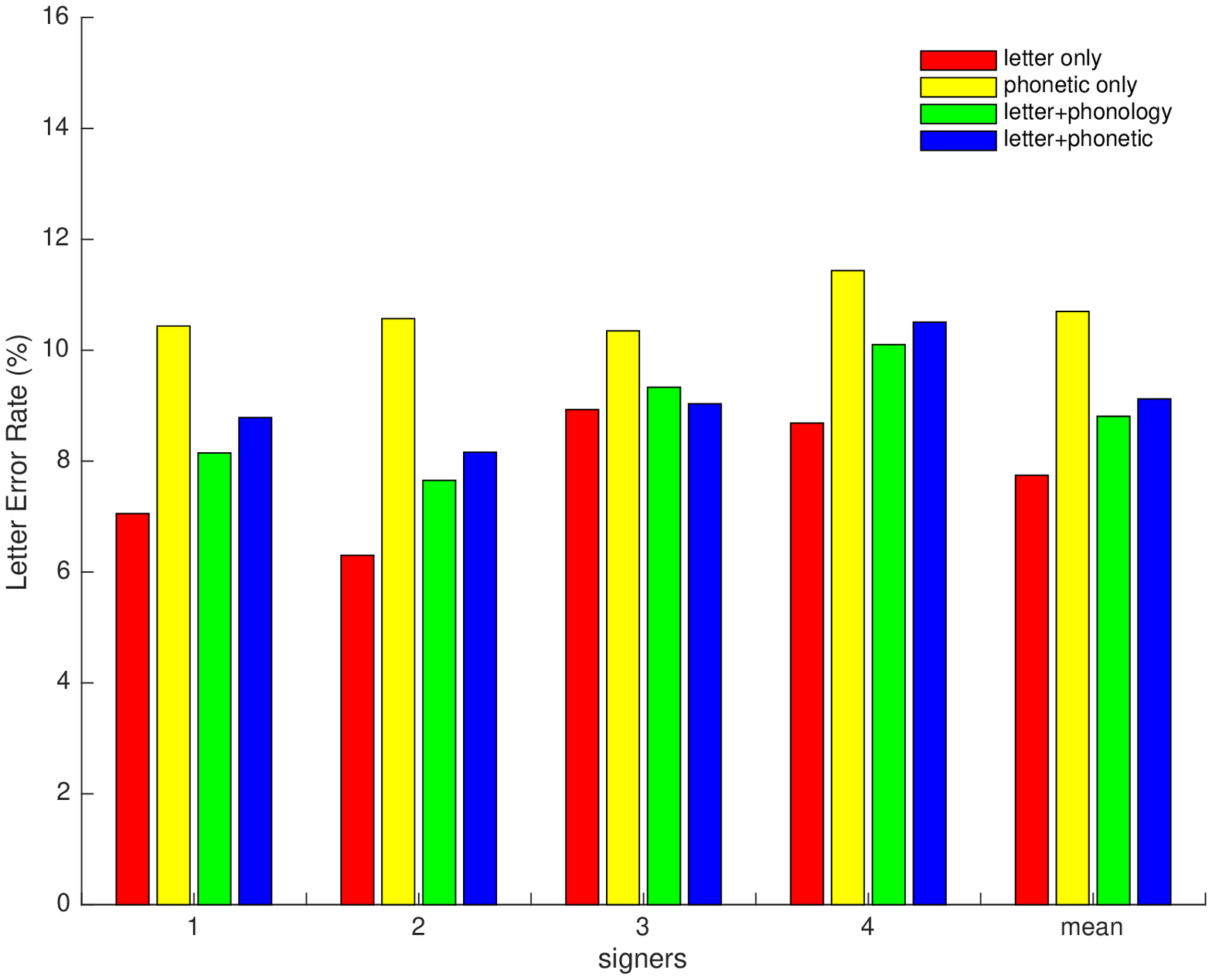}\\
  \includegraphics[width=0.8\linewidth]{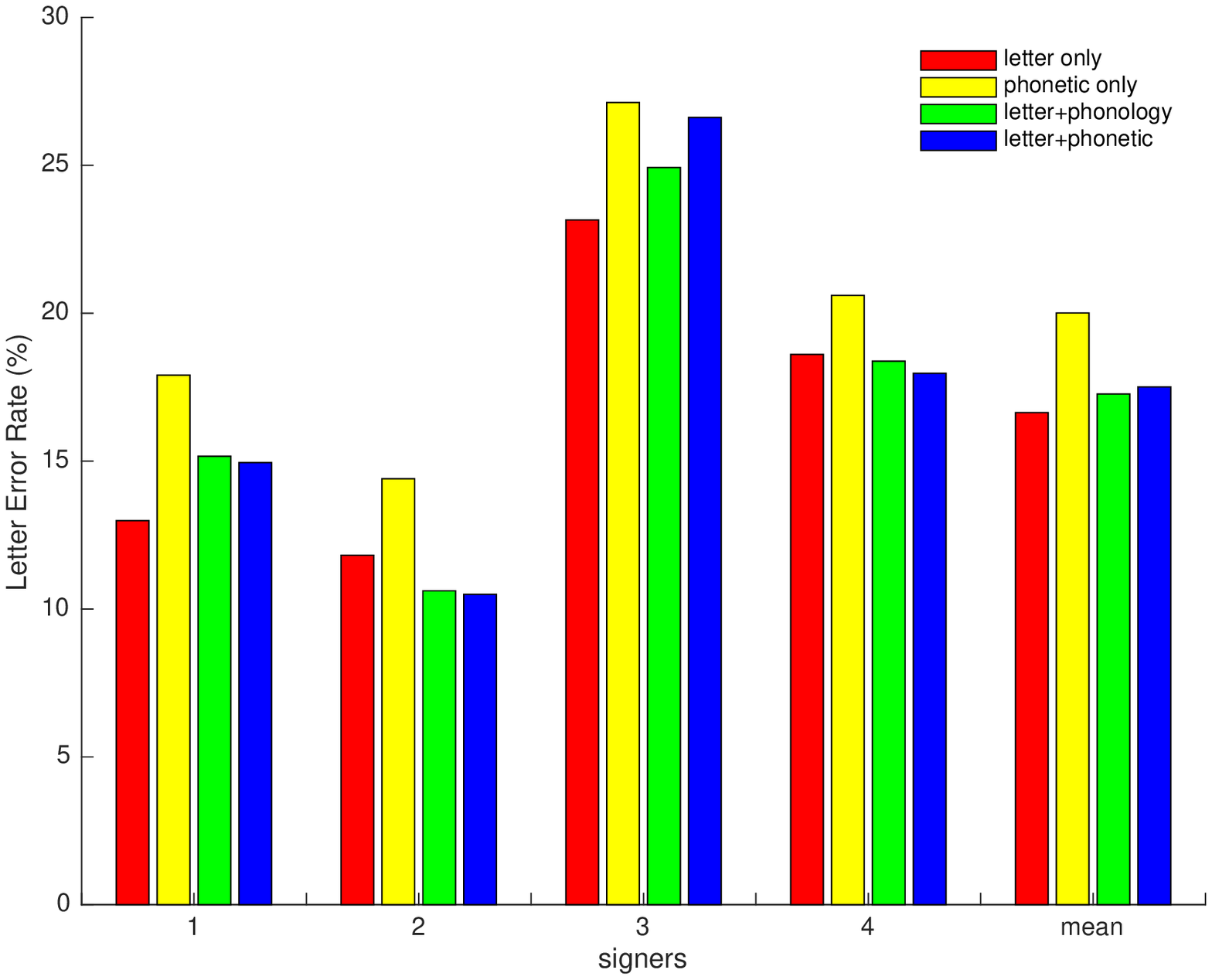} \\
\end{tabular}
\caption{(Top) Signer-dependent recognition and (bottom) signer-independent recognition with frame annotations and adaptation. We compare: FS-letter only, phonetic features only, FS-letters + phonological features \cite{bren} and FS-letters + phonetic features \cite{Keane2014diss}.} 
\label{fig:ling_fe}
\end{figure*}

We next compare the FS-letter DNN classifiers and the phonological
feature DNN classifiers in the context of the first-pass SCRF
recognizers.  We also consider an alternative sub-letter feature set,
in particular a set of phonetic features introduced by Keane \cite{Keane2014diss}, whose feature values are listed in Section~\ref{sec:app}, Tab.~\ref{t:phonetic}.  We use the first-pass SCRF
with either only FS-letter classifiers, only phonetic feature
classifiers, FS-letter + phonological feature classifiers, and FS-letter +
phonetic feature classifiers.  We do not consider the case of
phonological features alone, because they are not discrimative for
some FS-letters. Fig.~\ref{fig:ling_fe} shows the FS-letter recognition results
for the signer-dependent and signer-adapted settings.  

We find that using FS-letter classifiers alone outperforms the other
options in the signer-dependent setting, achieving 7.7\% letter error rate.  For signer-adapted
recognition, phonological or phonetic features are helpful in addition
to FS-letters for two of the signers (signers 2 and 4) but not for the
other two (signers 1,3); on average, using FS-letter classifiers alone is
best in both cases, achieving 16.6\% accuracy on average.  In contrast, in earlier work~\cite{kim2012} we found that phonological features 
  outperform FS-letters in the tandem HMM.  However, those experiments
  used a tandem model with neural networks with
  a single hidden layer; we conjecture that with more layers, we are
  able to do a 
  better job at the more complicated task of FS-letter classification.

\subsubsection{Analysis:  DNNs vs.~CNNs}
\label{sec:cnn-exp}

In all experiments thus far, we have used HOG image descriptors fed into fully connected feedforward DNNs. However, it has recently become common to use convolutional neural networks (CNNs) on raw image pixels without any hand-crafted image descriptors, which produces improved performance for certain visual recognition tasks (e.g., \cite{krizhevsky2012imagenet, simonyan2014very}).  In our case, we have relatively little training data compared with typical benchmark visual recognition tasks, motivating our choice to preprocess with image descriptors rather than rely on networks that learn features from raw pixels.  To test this assumption, we compare the performance of FS-letter and phonological feature classifiers based on DNNs and CNNs.  We use a signer-dependent setting and the same 8-fold setup.

For CNN experiments, our inputs are grayscale $64 \times 64 \times T$ pixels. $T$ is the number of frames used in the input window, as in our DNNs.  The CNNs use 32 kernels of $3 \times 3$ filters with stride 1 for the first and second convolutional layers. Next we add a max pooling layer with a $2 \times 2$ pixel window, with stride 2. For the third and fourth convolutional layer, we use 64 kernels of $3 \times 3$ filters with stride 1. Again, we then add a max pooling layer with a $2 \times 2$ pixel window, with stride 2. For all convolutional layers, ReLUs are used for the nonlinearity (as in our DNNs). Finally, two fully connected layers with 2000 ReLUs are added, followed by a softmax output layer.  We use dropout to prevent overfitting, with a 0.25 dropout probability for all convolutional layers and 0.5 for fully connected layers. Training is done via stochastic gradient descent with learning rate 0.01, learning rate decay $1\times10^{-6}$, momentum 0.9, and miini-batches of size 100. The network structure and all other settings were tuned on held-out data in preliminary experiments.  $T$ was also tuned and set to 21.  We implemented the CNNs using Keras \cite{chollet2015} with Theano \cite{2016arXiv160502688short}.

\begin{figure}\centering
\centering
\vspace{-.25in}
\hspace{-.0in}\includegraphics[width=0.9\linewidth]{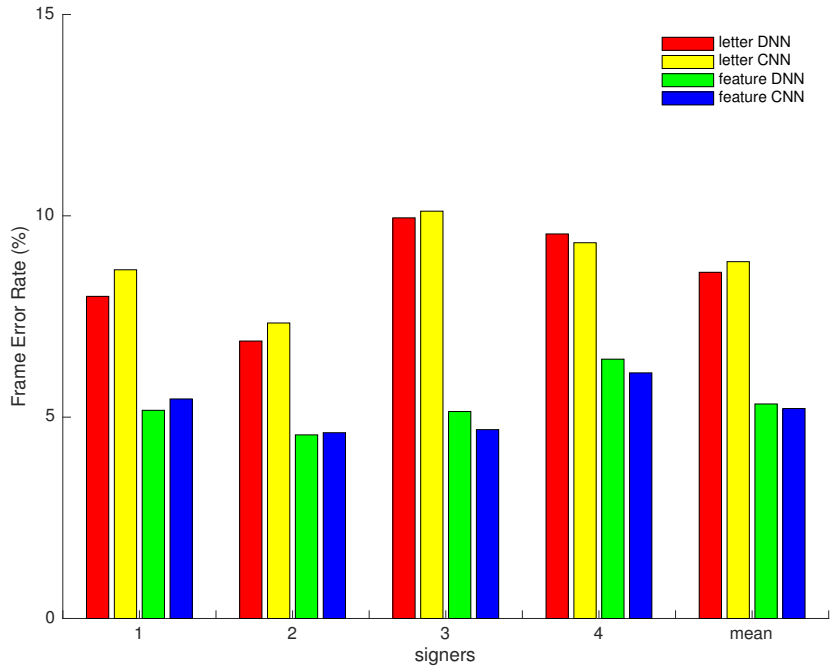} \\
\vspace{-.05in}
\caption[Comparison between DNN and CNN frame classifiers]{Comparison between DNN and CNN frame classifiers, in signer-dependent experiments over four signers.}
\label{fig:cnn_fer}
\end{figure}

Fig.~\ref{fig:cnn_fer} shows the results for the signer-dependent setting. We find that the performances of the DNNs and CNNs are comparable.  Specifically, for FS-letter classification, DNNs are slightly better, while for phonological feature classification, CNNs are slightly better.  However, in both cases the gaps are very small.  These results provide evidence for our hypothesis that on this data set, CNNs on raw pixels do not provide a substantial benefit, and we do not use them in further experiments.

\subsubsection{Improving performance in the force-aligned adaptation case}

We next attempt to improve the performance of adaptation in the
absence of ground-truth (manually annotated) frame labels.  This is an important setting, since in practice it can be very difficult to obtain ground-truth labels at the frame level.  Using only
the FS-letter label sequence for the adaptation data, we use the
signer-independent tandem recognizer to get force-aligned frame
labels.  We then adapt (fine-tune) the DNNs using the force-aligned
adaptation data (as before in the FA case).  We then
re-align the test signer's adaptation data with the adapted
recognizer. Finally, we adapt the DNNs again with the re-aligned data.
Throughout this experiment, we do not change the recognizer but only
update the DNNs.  Using this iterative realignment approach, we are able to furthur improve
the recognition accuracy in the FA case by about 1.3\%, as shown in Tab.~\ref{t:LER_fa}.

\begin{table*}[ht!]
\centering
\resizebox{\linewidth}{!}{
\begin{tabular}{|c|c|c|c|c||c|c|c|c|c|}
\hline
\multicolumn{5}{|c||}{Fine-tuning with FA} &
\multicolumn{5}{c|}{Fine-tuning with FA + realignment} \\ \hline
Signer 1 & Signer 2 & Signer 3 & Signer 4 & {\bf Mean} & Signer 1 & Signer 2 & Signer 3 & Signer 4 & {\bf Mean} \\ \hline\hline
22.7 & 26.0 & 33.4&  34.8&   {\bf 29.2}&  21.9& 25.3&  30.5&  34.0& {\bf 27.9} \\ \hline
\end{tabular}
}
\vspace{-.1in}
\caption{Letter error rates (\%) with iterated forced-alignment (FA) adaptation.}
\label{t:LER_fa}
\end{table*}

\subsubsection{Improving performance with segmental cascades}

Finally, we consider whether we can improve upon the performance of
our best models, the first-pass SCRFs, by rescoring their results in a second pass with
more powerful features.  We follow the discriminative segmental
cascades (DSC) approach of \cite{tang2015}, where a simpler first-pass SCRF is used for lattice
generation and a second SCRF, with more computationally demanding
features, is used for rescoring.  

For these experiments we start with the most successful first-pass
SCRF in the above experiments, which uses FS-letter DNNs and is adapted
with 20\% of the test signer's data with ground-truth peak handshape labels.  For the second-pass SCRF, we use the
first-pass score as a feature, and add to it two more complex
features:  a segmental DNN, which takes as input an entire
hypothesized segment and produces posterior probabilities for all of
the FS-letter classes; and the ``peak detection'' feature described in
Section~\ref{sec:method}. We use the same bigram language model as in the tandem HMM and rescoring SCRF models. For the segmental DNN, the training segments are given by the ground-truth segmentations derived from manual peak annotations.  The input layer consists of a concatenation of three mean HOG vectors, each averaged over one third of the segment.  We use the same DNN structure and learning strategy as for the DNN frame classifiers.

As shown in Tab.~\ref{t:DSC}, this approach
slightly improves the average FS-letter accuracy over four signers, from
16.6\% in the first pass to 16.2\% in the second pass.  This
improvement, while small, is statistically significant at the p=0.05
level (using the MAPSSWE significance test from the NIST Scoring Toolkit \cite{sctk}). These results combine our most successful ideas and form our final best results for signer-adapted recognition.  For the signer-dependent setting, this approach achieves comparable performance to the first-pass SCRF.

\begin{table*}[ht!]
\centering
\resizebox{\linewidth}{!}{
\begin{tabular}{|c|c|c|c|c||c|c|c|c|c|}
\hline
\multicolumn{5}{|c||}{Signer-dependent} &
\multicolumn{5}{c|}{Signer-independent} \\ \hline
Signer 1 & Signer 2 & Signer 3 & Signer 4 & {\bf Mean} & Signer 1 & Signer 2 & Signer 3 & Signer 4 & {\bf Mean} \\ \hline\hline
7.2& 6.5& 8.1&8.6&    {\bf 7.6}&  13.0& 11.2&  21.7&  18.8& {\bf 16.2} \\ \hline
\end{tabular}
}
\vspace{-.1in}
\caption{Letter error rates (\%) obtained with a two-pass segmental cascade.}
\label{t:DSC}
\end{table*}

\section{Conclusion}
This paper tackles the problem of unconstrained fingerspelled letter
sequence recognition in ASL, where the FS-letter sequences are not
restricted to any closed vocabulary.  This problem is challenging due
to both the small amount of available training data and the
significant variation between signers.  Our data
collection effort has thus far produced a set of carefully
annotated fingerspelled
letter sequences for four native ASL signers.  Our recognition
experiments have compared HMM-based and segmental models with features
based on DNN
classifiers, and have investigated a range of settings including
signer-dependent, signer-independent, and signer-adapted.  Our main contributions are:

\begin{itemize}
\item
  We have developed an approach for
    quick annotation of fingerspelling data using a two-pass approach,
    where multiple non-expert annotators quickly find candidate peak
    times, which are then automatically combined and verified by an
    ASL expert.

  \item Signer-dependent recognition, even with only a small amount of training data, is quite successful, reaching letter
    error rates below 10\%.  Signer-independent recognition, in contrast,
    is quite challenging, with letter error rates around 60\%.
    DNN adaptation allows us to bridge a
    large part of the gap between signer-independent and
    signer-dependent performance.

  \item Our best results are obtained using two-pass discriminative segmental
    cascades with features based on frame-level DNN FS-letter
    classifiers.  This approach achieves an average letter error rate
    of 7.6\% in the signer-dependent
    setting and 16.2\% LER in the signer-adapted setting.  The
    adapted models use signer-independent SCRFs and DNNs adapted by
    fine-tuning on the adaptation data.  The adapted 
    results are obtained using about 115 words of adaptation data with
    manual annotations of FS-letter peaks.

  \item If an even smaller amount of adaptation data is available, we can still get

    improvements from adaptation down to about 30 words of adaptation
    data.

  \item  In the absence of manual frame-level annotations, we can
    automatically align the adaptation data and still get a
    significant boost in performance from adaptation.  We can also iteratively improve the performance by re-aligning the
    data with the adapted models.  Our best adapted models using
    automatically aligned adaptation data achieve 27.3\% LER.

  \item The main types of errors that are addressed by adapting the
    DNN classifiers are confusions between signing and non-signing segments.
\end{itemize}

We are continuing to investigate additional sequence recognition
approaches as well as adaptation methods, especially for the case
where no manual annotations are available.
Future work will also expand our data collection to a larger number of
signers and to data collected ``in the wild'', such as
videos from Deaf online media.  Finally, future work will
consider jointly detecting and recognizing fingerspelling sequences
embedded within running ASL.

\section*{Acknowledgements}
We are grateful for the work of several undergraduate assistants for
annotation of the data.  This research was funded by a Google Faculty Award and 
by NSF grants NSF-1433485 and NSF/BCS-1251807.  The opinions expressed in this work are those of the authors and do not necessarily reflect the views of the
funding agency.

\appendix
\section{Fingerspelled handshape definitions and annotation
  conventions}
\label{sec:app}

\textbf{Handshape} -- We defined a handshape as stable if all of digits assumed a position and maintained it with only minor fluctuations. As soon as any digit moves, the handshape is considered to not be stable anymore. We were conservative with respect to holds, in that if a digit moves a small amount, but that movement is part of a larger movement that preceded or followed, that was not considered stable.
\begin{itemize}
  \item \textbf{Fingerspelled (FS)-letters} are defined as the target, canonical handshapes that are linked to the 26 letters of the English alphabet.
  \item \textbf{Peak handshapes} are the phonetically realized handshapes used in a given sequence at the moment
    when it most closely approximates the FS-letter.
\end{itemize}

\textbf{Orientation} -- Most \textsc{fs}-letters are produced with the palm facing away from the signer's body. The few exceptions to this are \fslist{g,h,p,q}\footnote{These are the \textsc{fs}-letters traditionally described as having different orientation; there are other possibilities that we have found as well: \fslist{x,y}.} where the palm faces the signer (\fslist{g,h}; labeled \emph{side} in our analysis), or faces down (\fslist{p,q}; labeled \emph{down} in our analysis). Because handshape and orientation changes are not always synchronized, we have annotated handshape stability as a hold, even if the hand is continuing to undergo an orientation change. Future annotation is necessary for orientation changes in detail and determine the pattern of stability and motion that exists there.

\textbf{Movement} -- Two \textsc{fs}-letters are described as having movement: \fslist{j,z}. \fslist{j} involves an orientation change, and \fslist{z} traces the path of the letter\footnote{This is frequently abbreviated to just a horizontal line, representing the top bar of the z.}. For both of these \textsc{fs}-letters, again we have annotated a hold to be where the handshape is stable, regardless of orientation change, or path movement.

\textbf{Handshape detail} -- A detailed (although not exhaustive) description of handshapes is given in table \ref{tab:handshape}. This is meant to be guidance to annotators, and is intended to catch the core features for each handshape, allowing for the systematic variation known to exist in handshape. If some of these features match, but the handshape is significantly different than expected, annotators added a diacritic ($+$) to note a large amount of deviance. This is not intended to exhaustively mark all of the deviant handshapes, but only those that should be looked into further. There are some instances where a peak is found, but no peak was detected. Although we have not analyzed this systematically, these instances are frequently peak handshapes that are instantaneous, or peak handshapes that occur extremely close to each other. These peaks are noted with a different diacritic (*). Finally if two handshapes have compressed to form a single peak handshape, a digraph is used to annotate the combined peak handshape. Examples that we have seen so far are \fslist{gh, it, in, io, il, ci}. Here the digraph is simply two FS-letters that seem to make up the single peak; for consistency they should be written in alphabetical order regardless of the orthographic order of the letters in the word being fingerspelled. See table \ref{tab:digraphs} for a description of those found so far.

\begin{table}
  \vspace{-1cm}
  \scriptsize
\begin{tabular}{lp{4in}}\toprule
\textsc{fs}-Letter & description of the most canonical shape\\\midrule
\fslist{a} & all fingers are flexed, with thumb touching the radial side of the hand, or extended\\ 
\fslist{b} & all fingers are extended. The thumb is hyper flexed across the palm\\ 
\fslist{c} & all the fingers are curved.\\ 
\fslist{d} & the index finger is fully extended. At least the middle finger is making contact with the thumb: the ring and pinky may be either flexed, or making contact with the thumb\\ 
\fslist{e} & the thumb is bent and hyper flexed across the palm, the index finger is bent and may be touching the thumb. The other fingers may be bent, like the index finger, or flexed completely.\\ 
\fslist{f} & contact with the index and thumb. The middle, ring, and pinky are all extended\\ 
\fslist{g} & the index finger is fully extended. All other fingers are flexed. The thumb is either extended fully, or unextended, against the middle finger.\\ 
\fslist{h} & index and middle fingers are fully extended. All others are flexed. The thumb is unextended or extended\\ 
\fslist{i} & the pinky is fully extended, all other fingers are flexed. The thumb is either hyperflexed, or unexetended, against the radial side\\ 
\fslist{j} & the pinky is fully extended, all other fingers are flexed. The thumb is either hyperflexed, or unexetended, against the radial side\\ 
\fslist{k} & the index finger is fully extended, the middle finger is extended, but bent ~90º at the joint closest to the hand\\ 
\fslist{l} & the index finger is fully extended, and the thumb is extended away from the hand. All other fingers are flexed\\ 
\fslist{m} & the index, middle, and ring fingers are closed, or flat-closed over the thumb, which is hyper flexed across the palm, possibly touching the base of the pinky and ring fingers\\ 
\fslist{n} & the index and middle fingers are closed, or flat-closed over the thumb, which is hyper flexed across the palm, possibly touching the base of the ring and middle fingers\\ 
\fslist{o} & the thumb and the index finger are touching in a curved, closed configuration. The other fingers are either in the same configuration, touching the thumb, or completely flexed.\\ 
\fslist{p} & the index finger is fully extended, the middle finger is extended, but bent ~90º at the joint closest to the hand\\ 
\fslist{q} & the index finger is fully extended. All other fingers are flexed. The thumb is either extended fully, or unextended, against the middle finger.\\ 
\fslist{r} & the index and middle fingers are extended and crossed over each other. All other fingers are flexed\\ 
\fslist{s} & all fingers are completely flexed, with thumb hyperflexed across the fist\\ 
\fslist{t} & the index finger is closed, or flat-closed over the thumb, which is hyper flexed across the palm, possibly touching the base of the middle and index fingers\footnote{There are some instances where the index finger flexes as the thumb is moving away from the base between the middle and index finger. In these cases the apogee for \fslist{t} should be marked when the index finger has started coming down, and the thumb starts moving. Frequently there looks to be a slight brush of the tip of the thumb across the proximal phalange.}\\ 
\fslist{u} & the index and middle fingers are completely extended, all other fingers are flexed, the thumb is hyperflexed across the palm\\ 
\fslist{v} & the index and middle fingers are completely extended and are spread apart, all other fingers are flexed, the thumb is hyperflexed across the palm\\ 
\fslist{w} & the index, middle, and ring fingers are completely extended and are spread apart, all other fingers are flexed, the thumb is hyperflexed across the palm\\ 
\fslist{x} & the index finger is bent similar to \fslist{e}, all other fingers are completely flexed\\ 
\fslist{y} & the pinky is fully extended, the thumb is hyper extended away from the hand. All other fingers are flexed\\ 
\fslist{z} & the index finger is fully extended, and all other fingers are flexed\\\bottomrule
\end{tabular}

\normalsize
\caption{Description of handshapes.}
\label{tab:handshape}
\end{table}

\begin{table}
  \vspace{-1cm}
  \footnotesize
\begin{tabular}{lp{4in}}\toprule
FS-letter & description of the most canonical handshape\\\midrule
\fslist{it} & the index finger is closed, or flat-closed over the thumb, which is hyper flexed across the palm, possibly touching the base of the middle and index fingers, and the pinky is extended. This should only be used if the hand reaches this configuration at a single frame.\\ 
\fslist{gh} & index and middle fingers as well as the thumb are extended. Similar to the \textsc{cl} 3 handshape.\\
\fslist{in} & index and middle fingers are (partially) flexed over the thumb, and the pinky is fully extended.\\
\fslist{io} & index, middle, and ring fingers are looped and touching the thumb, and the pinky is fully extended.\\
\fslist{il} & the index and thumb are extended (the thumb is abducted), and the pinky is fully extended.\\
\fslist{ci} & the index, middle, and ring fingers as well as the thumb are partially extended (also described as curved open), and the  pinky is fully extended. \\
\bottomrule
\end{tabular}
\normalsize
\caption{Description of digraphs.}
\label{tab:digraphs}
\end{table}

\section{Phonetic feature definitions}
\label{sec:phonfeat}

\begin{table*}[ht!]
\centering
\resizebox{\linewidth}{!}{
\begin{tabular}{ccccccccccccccc}
\hline
letter&\multicolumn{2}{c|}{index} & \multicolumn{2}{c|}{middle} & \multicolumn{2}{c|}{ring}& \multicolumn{2}{c|}{pinky}&spread&\multicolumn{4}{|c|}{thumb}&palm \\ \hline

& MCP &PIP& MCP& PIP& MCP& PIP& MCP& PIP& & y& z& PIP& touch \\ \hline \hline
a& 90& 90& 90& 90& 90& 90& 90& 90& 0& 0& 90& 180& i& for\\
b& 180& 180& 180& 180& 180& 180& 180& 180& 0& -45& 90& 180& r& for\\ 
c& 180& 90& 180& 90& 180& 90& 180& 90& 0& 0& 0& 135& -& for\\
d& 180& 180& 90& 135& 90& 90& 90& 90& 0& 0& 45& 180& m& for\\
e& 135& 90& 135& 90& 135& 90& 135& 90& 0& -45& 0& 90& r& for\\
f& 90& 135& 180& 180& 180& 180& 180& 180& 1& 0& 45& 180& i& for\\
g& 180& 180& 90& 90& 90& 90& 90& 90& 0& 0& 90& 180& m& in\\
h& 180& 180& 180& 180& 90& 90& 90& 90& 0& -45& 90& 180& r& in\\
i& 90& 90& 90& 90& 90& 90& 180& 180& 0& -45& 90& 180& r& for\\
j& 90& 90& 90& 90& 90& 90& 180& 180& 0& -45& 90& 180& r& dwn\\
k& 180& 180& 90& 180& 90& 90& 90& 90& 0& 0& 90& 180& m& for\\
l& 180& 180& 90& 90& 90& 90& 90& 90& 0& 90& 0& 180& -& for\\
m& 90& 135& 90& 135& 90& 135& 90& 90& 0& -45& 90& 180& p& for\\
n& 90& 135& 90& 135& 90& 90& 90& 90& 0& -45& 90& 180& r& for\\
o& 135& 135& 135& 135& 135& 135& 135& 135& 0& -45& 0& 180& m/i& for\\
p& 180& 180& 90& 180& 90& 90& 90& 90& 0& 0& 90& 180& m& dwn\\
q& 180& 180& 90& 90& 90& 90& 90& 90& 0& 0& 90& 180& m& dwn\\
r& 180& 180& 180& 180& 90& 90& 90& 90& -1& -45& 0& 180& r& for\\
s& 90& 90& 90& 90& 90& 90& 90& 90& 0& -45& 45& 180& r& for\\
t& 90& 135& 90& 90& 90& 90& 90& 90& 0& -45& 90& 180& m& for\\
u& 180& 180& 180& 180& 90& 90& 90& 90& 0& -45& 90& 180& r& for\\
v& 180& 180& 180& 180& 90& 90& 90& 90& 1& -45& 90& 180& r& for\\
w& 180& 180& 180& 180& 180& 180& 90& 90& 1& -45& 90& 180& p& for\\
x& 180& 135& 90& 90& 90& 90& 90& 90& 0& -45& 45& 180& m& for\\
y& 90& 90& 90& 90& 90& 90& 180& 180& 1& 90& 0& 180& -& for\\
z& 180& 180& 90& 90& 90& 90& 90& 90& 0& 0& 45& 180& m& for\\
zz& 180& 180& 180& 180& 90& 90& 90& 90& 1& 0& 45& 180& m& for\\ \hline 
\end{tabular}
}
\vspace{-.1in}
\caption{Phonetic features \cite{Keane2014diss}. The numerical values
  refer to joint angles in each finger.}
\label{t:phonetic}
\end{table*}

\clearpage

\section*{References}

\bibliography{description,csl2016}

\end{document}